\documentclass[a4paper,fleqn]{cas-dc}

\usepackage[authoryear,longnamesfirst]{natbib}

\usepackage{graphicx}
\usepackage{amsmath,amssymb,amsfonts}
\usepackage{xcolor}
\usepackage{textcomp}
\usepackage{booktabs}
\usepackage{algorithm}
\usepackage{algorithmicx}
\usepackage{algpseudocode}
\usepackage{listings}
\usepackage{placeins}
\usepackage{tabularx}
\usepackage{float}

\makeatletter
\ExplSyntaxOn
\newcommand{\figcaption}[1]{%
  \def\@captype{figure}%
  \dim_set:Nn \l_fig_width_dim { \linewidth }%
  \tl_set:Nn \l_fig_align_tl { \centering }%
  \cs_set_eq:NN \@makecaption \__make_fig_caption:nn
  \caption{#1}%
}
\ExplSyntaxOff
\makeatother

\newdefinition{definition}{Definition}

\begin{document}
\let\WriteBookmarks\relax
\def\floatpagepagefraction{1}
\def\textpagefraction{.001}

\shorttitle{Hybrid Document-Routed Retrieval for Financial RAG}
\shortauthors{Z.\ Cheng et~al.}

\title[mode=title]{Sustainable Hybrid Document-Routed Retrieval for Financial RAG: Resolving the Robustness-Precision Trade-off}

\author[1]{Zhiyuan Cheng}[orcid={0009-0003-2845-5322}]
\cormark[1]
\ead{zhycheng@stanford.edu}
\credit{Conceptualization, Methodology, Software, Writing -- Original draft, Visualization}

\affiliation[1]{organization={School of Engineering, Stanford University},
    city={Stanford},
    state={CA},
    country={USA}}

\author[2]{Longying Lai}
\credit{Data curation, Validation, Writing -- Review \& Editing}

\affiliation[2]{organization={Simon Business School, University of Rochester},
    city={Rochester},
    state={NY},
    country={USA}}

\author[3]{Yue Liu}
\credit{Validation, Writing -- Review \& Editing}

\affiliation[3]{organization={Accounting \& Information Systems, Rutgers University},
    city={Newark},
    state={NJ},
    country={USA}}

\cortext[1]{Corresponding author}

\begin{abstract}
Retrieval-Augmented Generation (RAG) systems for financial document question answering typically follow a chunk-based paradigm: documents are split into fragments, embedded into vector space, and retrieved via similarity search. While effective in general settings, this approach suffers from cross-document chunk confusion in structurally homogeneous corpora such as regulatory filings. Semantic File Routing (SFR), which uses LLM structured output to route queries to whole documents, reduces catastrophic failures but sacrifices the precision of targeted chunk retrieval. We identify this robustness-precision trade-off through controlled evaluation on the FinDER benchmark (1,500 queries across five groups): SFR achieves higher average scores (6.45 vs.\ 6.02) and fewer failures (10.3\% vs.\ 22.5\%), while chunk-based retrieval (CBR) yields more perfect answers (13.8\% vs.\ 8.5\%). To resolve this trade-off, we propose \emph{Hybrid Document-Routed Retrieval} (HDRR), a two-stage architecture that uses SFR as a document filter followed by chunk-based retrieval scoped to the identified document(s). HDRR eliminates cross-document confusion while preserving targeted chunk precision. Experimental results demonstrate that HDRR achieves the best performance on every metric: an average score of 7.54 (25.2\% above CBR, 16.9\% above SFR), a failure rate of only 6.4\%, a correctness rate of 67.7\% ($+$18.7~pp over CBR), and a perfect-answer rate of 20.1\% ($+$6.3~pp over CBR, $+$11.6~pp over SFR). HDRR resolves the trade-off by simultaneously achieving the lowest failure rate and the highest precision across all five experimental groups. Beyond accuracy, HDRR is also the most efficient of the high-quality systems: it preserves CBR's compact per-query token budget ($\sim$5K--15K, an order of magnitude below SFR's full-document $\sim$50K--200K), incurs no indexing-time LLM spend (in contrast to the one-time $\sim$\$100 cost of LLM-context contextual indexing), and uses fewer per-query LLM calls than self-correcting agentic baselines---translating directly to lower API spend and lower inference-time energy at deployment scale.
\end{abstract}

\begin{keywords}
Retrieval-Augmented Generation \sep Financial Document Analysis \sep Question Answering \sep Hybrid Retrieval \sep Document Routing \sep Sustainable AI \sep Green AI \sep 10-K Reports
\end{keywords}

\maketitle

\section{Introduction}\label{sec:introduction}

\subsection{Background and Motivation}

Financial 10-K reports are comprehensive annual filings mandated by the U.S.\ Securities and Exchange Commission (SEC) for all publicly traded companies. A typical 10-K report for a large corporation spans 100--300 pages and encompasses audited financial statements, management discussion and analysis (MD\&A), risk factor disclosures, and supplementary schedules. For the S\&P~500 index alone, these filings collectively represent tens of thousands of pages of dense, domain-specific text that analysts, investors, and regulators must navigate to extract actionable information. Recent advances in generative AI have broadened the scope of automated analysis across such complex, real-world domains~\citep{yao2025generative}, ranging from urban traffic-signal control~\citep{deng2026adaptive} to deployment frameworks for AI in healthcare~\citep{chen2026applying}.

Retrieval-Augmented Generation (RAG)~\citep{lewis2020rag} has emerged as a leading paradigm for building question-answering (QA) systems over such large document collections. The standard RAG pipeline operates in two phases: an offline \emph{indexing} phase that splits documents into chunks, computes embeddings, and stores them in a vector database; and an online \emph{query} phase that retrieves the most relevant chunks for a given question and feeds them as context to a large language model (LLM) for answer generation. This chunk-based approach has proven effective across many domains~\citep{gao2023retrieval, izacard2022atlas, siriwardhana2023improving}.

However, applying chunk-based RAG to corpora of structurally homogeneous documents reveals a class of failure modes that are underexplored in the literature. Financial regulatory filings such as 10-K reports share standardized section headings, boilerplate language, and tabular formats mandated by SEC regulations. When hundreds of such documents are chunked and indexed together, structurally similar sections from \emph{different} companies become nearly indistinguishable in embedding space. A query about ``Apple's revenue recognition policy'' may retrieve chunks from Microsoft's or Google's 10-K report because all three companies use similar regulatory language in their revenue recognition disclosures. We term this phenomenon \emph{cross-document chunk confusion}, and it represents a fundamental limitation of chunk-level retrieval in homogeneous corpora.

Existing remedies to this failure mode trade accuracy against efficiency in different ways. Routing the entire document to the generation model eliminates cross-document confusion but inflates per-query token budgets by an order of magnitude. Contextual indexing approaches~\citep{anthropic2024contextual} sharpen chunk-level discrimination at the cost of a per-chunk LLM-context-generation pass over the entire corpus---a non-trivial one-time spend. Self-correcting agentic designs~\citep{asai2024selfrag} pay multiple LLM calls per query for post-hoc verification and retry. Since LLM inference compute---and therefore deployment-time API spend and inference energy---scales with both per-query tokens and per-query LLM calls, an architecture that resolves the accuracy failure mode without paying these costs has direct economic and operational value at production scale. This concern aligns with the broader \emph{Green AI} agenda~\citep{schwartz2020green, strubell2019energy}, which calls for the inclusion of computational cost alongside accuracy as a first-class evaluation criterion; recent work~\citep{luccioni2024power} further establishes that inference---not just training---is a dominant contributor to AI's deployment-time energy footprint, making per-query token and per-query call counts directly relevant sustainability metrics for RAG systems.

\subsection{Proposed Approach}

This paper first introduces \textbf{Semantic File Routing (SFR)}, an alternative RAG paradigm designed for structurally regular document corpora. Instead of chunking and indexing documents, SFR leverages the ability of modern LLMs to produce structured output~\citep{openai2024structured} to \emph{resolve document identity} directly from the natural language query. Given a query, the LLM extracts document-identifying metadata (in our financial domain, the company ticker symbol and fiscal year) as a structured JSON object. These metadata fields are then mapped to a file path in a pre-organized repository (e.g., \texttt{\{year\}/\{ticker\}.pdf}), and the \emph{entire} document is provided as context to the generation model.

SFR eliminates cross-document confusion by construction, but our comparative evaluation reveals a fundamental trade-off: SFR reduces catastrophic failures while chunk-based retrieval (CBR) achieves higher precision when it retrieves correctly. To resolve this trade-off, we propose \textbf{Hybrid Document-Routed Retrieval (HDRR)}, a two-stage architecture that uses SFR's document routing as a first-stage filter, followed by chunk-based retrieval \emph{scoped to the identified document(s)}. HDRR eliminates cross-document chunk confusion (the primary source of CBR failures) while preserving the precision advantage of targeted chunk retrieval (which SFR's full-document context dilutes). With a graceful fallback to full-corpus search when routing fails, HDRR combines the robustness of document-level routing with the precision of chunk-level retrieval. Critically, HDRR achieves this without paying the token cost that other accuracy-focused designs incur: per-query generation context stays at the CBR level ($\sim$5K--15K tokens, an order of magnitude below SFR's full-document context), and the architecture requires no indexing-time LLM spend---making HDRR not only the most accurate but also the most efficient of the high-quality systems we compare against.

\subsection{Contributions}

This work makes the following contributions:

\begin{enumerate}
    \item \textbf{Trade-off Identification:} Through controlled evaluation of CBR and SFR on 1,500 queries across five independent groups from the FinDER benchmark~\citep{finder2025}, we identify and quantify a fundamental robustness-precision trade-off between chunk-level and document-level retrieval paradigms.

    \item \textbf{Hybrid Document-Routed Retrieval:} We propose HDRR, a two-stage architecture that uses LLM-based document routing to scope chunk-based retrieval, resolving the identified trade-off. HDRR achieves the best performance on all four evaluation metrics, simultaneously attaining the lowest failure rate and the highest answer precision.

    \item \textbf{Three-Paradigm Comparative Evaluation:} We conduct a rigorous three-way comparison of CBR, SFR, and HDRR using identical query sets, providing statistically robust evidence that the hybrid approach dominates both baseline paradigms.

    \item \textbf{Cost-Efficiency Analysis:} We quantify the per-query token budget, per-query LLM-call count, and indexing-time LLM spend of HDRR against five comparison systems (CBR, SFR, Agentic RAG, and two contextual-indexing variants), establishing that HDRR achieves the highest answer quality at zero indexing-time LLM cost and CBR-level per-query token cost---an order of magnitude below SFR. We further connect token cost to inference-time energy, framing HDRR as a cost-efficient deployment choice for production RAG systems.

    \item \textbf{Generalizability Analysis:} We formalize the corpus properties (naming regularity and structural homogeneity) that determine when document-level routing is beneficial, and discuss applicability to structured-document domains beyond finance.
\end{enumerate}

\subsection{Paper Organization}

The remainder of this paper is organized as follows. Section~\ref{sec:related_work} surveys related work on RAG systems, financial document analysis, and LLM structured output. Section~\ref{sec:problem} formulates the problem and defines the three retrieval paradigms. Section~\ref{sec:architectures} describes the architectures of all three systems in detail. Section~\ref{sec:experiments} presents the experimental setup and evaluation methodology. Section~\ref{sec:results} reports quantitative results. Section~\ref{sec:discussion} provides in-depth analysis and discussion. Section~\ref{sec:conclusion} concludes with a summary and future research directions.

\section{Related Work}\label{sec:related_work}

\subsection{Retrieval-Augmented Generation}

Retrieval-Augmented Generation (RAG), introduced by \citet{lewis2020rag}, augments language model generation with relevant passages retrieved from an external knowledge base. The core architecture comprises a retriever that identifies relevant documents given a query and a generator that produces responses conditioned on both the query and the retrieved context. This paradigm has become a standard approach for knowledge-intensive NLP tasks, with numerous variants proposed to improve retrieval quality, generation fidelity, and end-to-end performance~\citep{gao2023retrieval}.

A comprehensive survey by \citet{gao2023retrieval} categorizes RAG systems along three dimensions: indexing strategies (how documents are preprocessed and stored), retrieval strategies (how relevant passages are identified), and generation strategies (how the LLM produces answers from retrieved context). Our work contributes to this taxonomy by introducing a fourth dimension: \emph{retrieval granularity}, contrasting chunk-level retrieval with document-level routing.

\subsection{Hybrid Retrieval and Reranking}

Modern RAG systems frequently combine multiple retrieval strategies to improve recall. Hybrid search approaches merge sparse retrieval methods such as BM25~\citep{robertson2009bm25} or full-text search with dense semantic retrieval using learned embeddings~\citep{karpukhin2020dense, wang2024searching}. Reciprocal Rank Fusion (RRF)~\citep{cormack2009rrf} provides an unsupervised method for combining ranked lists from different retrievers, assigning each document a fused score based on its ranks across retrieval methods.

Neural reranking further refines retrieval quality by re-scoring candidates with cross-encoder models that jointly encode query-document pairs~\citep{nogueira2019passage, reimers2019sbert}. Cross-encoders capture fine-grained interaction patterns between queries and passages that bi-encoder models cannot, at the cost of increased computational overhead. Models such as Jina Reranker v2~\citep{jina2024reranker} and ColBERT~\citep{khattab2020colbert} represent the current state of the art in neural reranking. Our prior work~\citep{cheng2026reranking} demonstrated that neural reranking yields a 15.5 percentage point improvement in answer correctness for financial RAG, establishing it as a critical component of the chunk-based pipeline that serves as the baseline in this study.
The effectiveness of transformer-based dense embeddings for short-text retrieval has also been empirically examined across domains including e-commerce~\citep{transformerecommerce}, providing foundational validation for embedding-based chunk retrieval in specialized corpora. The principle of refining a coarse first-stage match into a precise final result recurs across modalities: composed image and video retrieval systems explicitly parse fine-grained modification semantics~\citep{li2025finecir} and bind query entities to candidate content~\citep{li2025encoder} as a second-stage refinement, mirroring the retrieve-then-refine structure that motivates reranking in text RAG.

\subsection{Financial Document Analysis}

Financial document analysis has attracted significant attention from both the NLP and financial technology communities~\citep{chen2024financial, loukas2023making}. The FinDER dataset~\citep{finder2025}, introduced by LinqAlpha, provides 5,703 query-evidence-answer triplets derived from real-world financial inquiries across 10-K filings. The dataset reflects realistic analyst workflows with ambiguous, concise queries featuring domain-specific abbreviations and jargon. Its characteristics, including 84.5\% qualitative questions, 15.5\% quantitative questions (with half requiring multi-step reasoning), and evidence often scattered across multiple document sections, make it a challenging benchmark for RAG systems.

Prior work on FinDER reports that advanced models such as GPT-4-Turbo achieve only 9\% accuracy on closed-book questions, demonstrating the necessity of retrieval-augmented approaches~\citep{finder2025}. Complementing FinDER, QuantEval~\citep{kang2026quanteval} provides a benchmark evaluating LLMs across three dimensions of quantitative finance---knowledge-based QA, mathematical reasoning, and strategy coding---revealing substantial gaps between frontier models and human experts particularly in reasoning tasks. On the generative side, FinSentLLM~\citep{zhang2025finsentllm} demonstrates that multi-LLM ensemble approaches can achieve consistent gains in financial sentiment analysis, with structured semantic signals providing econometric evidence linking model predictions to market dynamics. Multi-hop reasoning in RAG systems has been explored through graph-based and tree-based retrieval~\citep{shi2026reasoning}, and context pruning techniques such as AttentionRAG~\citep{fang2025attentionrag} aim to select only the most relevant passages from retrieved candidates. Beyond retrieval-centric QA, a substantial body of work mines signals directly from 10-K narratives and market data: interpretable text scores extracted from filing narratives support bankruptcy prediction~\citep{zhang2026bankruptcy}, interpretable factor decomposition informs decision intelligence in equity markets~\citep{han2026interpretable}, calibrated technical-indicator strategies model index dynamics~\citep{lin2026volume}, and disclosure-driven studies examine analyst behavior and greenwashing~\citep{dai2023analyst} as well as the information content of alternative data such as satellite imagery in cross-listed share pricing~\citep{dai2023neighbors}. These efforts underscore the analytical value locked in financial filings and motivate retrieval systems---like the ones studied here---that can surface the right document content reliably.

\subsection{Contextual Retrieval}\label{sec:rw_contextual}

A direct line of attack on cross-document chunk confusion is to enrich each chunk's representation with document-level context before embedding. \emph{Contextual Retrieval}~\citep{anthropic2024contextual} prepends an LLM-generated 50--100 token paragraph describing each chunk's role within its parent document (company, fiscal year, section, topic) prior to indexing, exploiting prompt caching of the full document to keep per-chunk generation costs low. The expectation is that the contextualized embedding distinguishes structurally identical sections across different filings.

A simpler ablation of the same idea prepends only a static \texttt{[Company: TICKER, Fiscal Year: YEAR]} tag to each chunk. Together, these two variants---a low-cost static prefix and the faithful LLM-generated context---form a natural pair of contextual baselines that probe the limits of in-chunk identity tagging. We include both as comparison systems in Sections~\ref{sec:experiments} and~\ref{sec:results}; our results confirm that contextual indexing meaningfully narrows the failure gap but does not close it on a structurally homogeneous corpus, because residual semantic overlap across regulatory filings remains in the embedding space after tagging.

\subsection{Knowledge-Graph and Global-Synthesis RAG}\label{sec:rw_graphrag}

A parallel line of work augments retrieval with a knowledge graph constructed over the corpus. \emph{GraphRAG}~\citep{edge2024graphrag} extracts entity-relationship triples from every document via repeated LLM calls during indexing, builds community summaries over the resulting graph, and answers queries by traversing communities rather than by retrieving raw passages. The framework's authors draw a sharp distinction between \emph{global} queries (questions that require synthesizing information across many documents, e.g., ``What are the common risk themes across the corpus?'') and \emph{local} queries (entity-specific lookups), and explicitly report that standard chunk RAG outperforms GraphRAG on the local-query regime.

Our setting is the local regime by construction: FinDER queries~\citep{finder2025} name a specific company and ask for a specific fact or policy from that company's filing. The within-document multi-section reasoning required by quantitative subqueries is a within-document retrieval problem (which HDRR's scoped chunk retrieval addresses by searching all chunks of the routed document, including non-adjacent sections), not a cross-document graph-traversal problem. Furthermore, graph construction over a 500-document, tens-of-thousands-of-pages corpus is dominated by LLM extraction calls and is orders of magnitude more expensive than the chunking-and-embedding indexing used by all systems in this study. For these reasons we discuss GraphRAG as a complementary paradigm for the global-query regime rather than a comparison baseline; the cost-and-scope argument is revisited in Section~\ref{sec:cost}.

\subsection{Self-Correcting and Agentic Retrieval}\label{sec:rw_agentic}

A third line of work replaces upfront retrieval-side improvements with a self-correcting agent loop that retrieves, inspects its own retrieval, and decides whether to refine. \emph{Self-RAG}~\citep{asai2024selfrag} trains an LLM to emit reflection tokens that signal whether to retrieve, whether retrieved passages are relevant, and whether the generated answer is supported; \emph{Reflexion}~\citep{shinn2023reflexion} extends the same principle to a general verbal-reinforcement loop. In production RAG systems, the equivalent pattern is a thin verification call after each retrieval that re-runs the retrieval with an entity-anchored query when verification fails. We refer to this family as \emph{Agentic RAG}.

For structurally homogeneous corpora, the conceptual contrast is clear: an Agentic RAG agent corrects wrong-company retrievals \emph{after the fact}, whereas the hybrid architecture proposed in this paper avoids them \emph{by construction} via upfront routing. We implement a minimal Agentic RAG baseline as a thin wrapper around our existing chunk-retrieval pipeline (one verification call plus up to two entity-anchored retries) and evaluate it alongside the routing-based systems. Section~\ref{sec:results} reports the empirical cost: Agentic's variable per-query call count is higher than HDRR's deterministic single routing call, while its verification step is structurally incompatible with multi-document queries because its ``majority-of-chunks-from-target'' criterion votes against legitimate cross-document retrievals. The broader problem of detecting and resolving when retrieved context conflicts with or contradicts the query has motivated a growing line of work on knowledge-conflict diagnosis~\citep{chen2026doesrag, ye2026seeing} and on calibrating a system's reliance on retrieved evidence~\citep{qian2026relevant}; lightweight small-model agents have also been proposed to keep such verification affordable at scale~\citep{zhang2026smalllanguage}. HDRR sidesteps the wrong-document instance of this problem upstream by scoping retrieval to the routed document, but these post-hoc mechanisms remain complementary for residual within-document conflicts.

\subsection{LLM Structured Output}

A recent development in LLM capabilities is the ability to generate outputs conforming to a specified schema, commonly referred to as \emph{structured output} or \emph{constrained decoding}~\citep{openai2024structured}. By providing a JSON schema, developers can instruct the model to produce outputs that are guaranteed to parse correctly, enabling reliable extraction of structured information from unstructured text.

Structured output has been applied to information extraction, data normalization, and tool use, but its application to retrieval, specifically to resolve document identity as a routing mechanism, has not been systematically studied. More broadly, prompt-engineered LLMs demonstrate strong few-shot classification capability across domain-specific tasks~\citep{liu2024poliprompt}, achieving reliable extraction without task-specific retraining---a property directly exploited by our SFR component. As frontier LLMs are increasingly deployed in structured real-world settings, their evaluation in realistic operational contexts has attracted growing attention~\citep{wang2026visualleak}, including diagnostic studies of where LLM-driven reward design fails under sparse structured signals~\citep{wang2026rewarddesign} and reward-free probes for prompted implicit hacking~\citep{shen2026selfcommitment}; our study contributes such an evaluation in the financial document QA domain, where structured output extraction directly determines retrieval quality. Our SFR approach leverages structured output as the core retrieval mechanism: the LLM extracts document-identifying metadata (ticker symbols and fiscal years) from natural language queries, effectively converting a retrieval problem into a structured classification problem.

\subsection{Long-Context LLMs and Document-Level Reasoning}

The emergence of LLMs with extended context windows (GPT-4 supports 128K tokens~\citep{openai2024gpt4}, and newer models push even further) has reopened the question of whether chunk-based retrieval is always necessary. \citet{liu2024lost} demonstrated that LLMs struggle to utilize information in the middle of long contexts (the ``lost in the middle'' phenomenon), while \citet{chen2023benchmarking} explored positional interpolation techniques for extending context windows.

\citet{jiang2024longrag} proposed LongRAG, which retrieves longer document segments rather than short chunks, demonstrating that coarser retrieval granularity can improve performance when combined with long-context models. Our SFR approach takes this concept to its logical extreme: retrieving the \emph{entire} document, relying on the LLM's long-context capabilities to process full financial filings (typically 50K--200K tokens after conversion).

\subsection{Document Routing and Classification}

Document routing, defined as the task of directing a query to the most relevant document or subset of documents, has been studied in the information retrieval literature, traditionally in the context of distributed search~\citep{cormack2009rrf}. In a distributed search system, a routing algorithm determines which document shards are most likely to contain relevant results, avoiding exhaustive search across all shards.

SFR adapts this concept by treating each document in the corpus as a ``shard'' and using LLM structured output as the routing function. Unlike traditional routing methods that rely on statistical models of shard contents, SFR exploits the LLM's world knowledge to map query entities (company names, ticker symbols) to file-system locations. This approach is most effective when the mapping between query entities and documents is predictable and consistent, a condition naturally satisfied by regulatory filing corpora. Routing mechanisms have also been studied in other AI-driven systems: for example, risk-aware dynamic routing in smart logistics~\citep{xue2026routing} similarly dispatches tasks to appropriate processing paths based on predicted network conditions, demonstrating the generality of query-driven routing as a systems design principle across domains.

\subsection{Multi-Stage and Cascaded Retrieval}

Multi-stage retrieval architectures decompose the retrieval process into successive refinement steps. The classic two-stage retrieve-then-rerank paradigm~\citep{nogueira2019passage} uses a fast first-stage retriever to generate candidates, followed by a computationally expensive cross-encoder reranker. More recent work extends this to three or more stages, incorporating document filtering, passage retrieval, and answer extraction as distinct pipeline steps.

In distributed information retrieval, resource selection algorithms first identify the most relevant document collections (``shards'') before performing within-shard retrieval~\citep{cormack2009rrf}. This shard-selection approach is conceptually related to our hybrid architecture, where document routing serves as a shard selector that restricts subsequent chunk retrieval to the identified document(s). However, traditional resource selection relies on statistical models of collection contents, whereas our approach leverages LLM structured output for zero-shot document identification based on query semantics.

The idea of combining document-level and chunk-level retrieval has been explored in enterprise search systems, but systematic evaluation of such hybrid architectures for RAG, particularly with LLM-based routing as the first stage, remains limited. Recent work in LLM-based agentic systems has explored difficulty-aware orchestration~\citep{su2026daao}, where predicted query complexity dynamically determines workflow structure and which model handles each step, as well as pattern-aware tool-integrated reasoning that learns \emph{how} to invoke retrieval and other tools rather than merely \emph{when}~\citep{xu2026learning}, supported by structured mid-level supervision for tool-using models~\citep{jiang2026scribe} and router-guided multi-teacher distillation for synthesizing the requisite training data~\citep{zhang2026optimalteacher}. A related strand cautions that the prompt-level design of such reasoning steps is delicate---corrective hints can backfire and induce overcaution~\citep{qi2026corrective}---and that auxiliary pipeline stages (e.g., coverage-aware crawling for domain-specific discovery~\citep{qi2026coverage} or logical self-reflection to guard against adversarial inputs~\citep{lin2026reflect}) carry their own reliability trade-offs. HDRR follows a related principle at the retrieval level: the routing stage classifies each query by whether document identity can be resolved, adapting the retrieval scope accordingly and falling back gracefully when routing confidence is low.

\subsection{Gap Analysis}

While individual components of our approach (RAG, structured output, long-context LLMs, document routing, contextual indexing, agentic retrieval, and multi-stage retrieval) have been studied independently, their integration into a coherent hybrid retrieval paradigm for structured-document corpora has not been explored. The closest paradigms---Contextual Retrieval~\citep{anthropic2024contextual} and Agentic RAG~\citep{asai2024selfrag}---attack chunk confusion either at indexing time (richer chunk representations) or at inference time (post-hoc verification and retry), but in both cases the chunk-level retrieval step still operates over the full corpus and therefore still admits wrong-document chunks into the candidate pool. Our work challenges the underlying assumption that chunk-level retrieval is the right primitive in this regime by: (1)~demonstrating that document-level routing outperforms chunk-level retrieval on aggregate metrics while introducing a robustness-precision trade-off, (2)~proposing a hybrid architecture that resolves this trade-off by combining document routing with scoped chunk retrieval, and (3)~empirically comparing the hybrid architecture against both the contextual-indexing and agentic-retrieval families on the same benchmark.

\section{Problem Formulation}\label{sec:problem}

\subsection{Task Definition}

We address the task of \emph{financial document question answering}: given a natural language query $q$ about a specific company's financial performance, operations, or disclosures, and a corpus of financial documents $\mathcal{D} = \{d_1, d_2, \ldots, d_N\}$, the system must produce a natural language answer $a$ that accurately and completely addresses the query based on information contained in $\mathcal{D}$.

In our setting, the corpus $\mathcal{D}$ consists of 10-K annual reports filed by S\&P~500 companies. Each document $d_i$ is associated with a company identifier (ticker symbol) $t_i$ and a fiscal year $y_i$, yielding a bijective mapping:
\begin{equation}\label{eq:doc_mapping}
    f: (t, y) \mapsto d \in \mathcal{D}
\end{equation}
where each (ticker, year) pair uniquely identifies a document. This structural regularity is a defining characteristic of the corpus and a prerequisite for the SFR paradigm.

\subsection{Retrieval Paradigms}

We formally define and compare three retrieval paradigms that differ in their retrieval unit, mechanism, and scope.

\subsubsection{Chunk-Based Retrieval (CBR)}

The chunk-based paradigm operates on sub-document units. During an offline indexing phase, each document $d_i$ is partitioned into overlapping text segments:
\begin{equation}
    d_i \rightarrow \{c_{i,1}, c_{i,2}, \ldots, c_{i,M_i}\}
\end{equation}
where $c_{i,j}$ denotes the $j$-th chunk of document $d_i$ and $M_i$ is the number of chunks. Each chunk is embedded into a vector space $\mathbb{R}^n$ via an embedding function $\phi$:
\begin{equation}
    \mathbf{v}_{i,j} = \phi(c_{i,j}) \in \mathbb{R}^n
\end{equation}

At query time, the system retrieves a set of top-$k$ chunks via similarity search:
\begin{equation}\label{eq:cbr_retrieval}
    \mathcal{R}_{\text{CBR}}(q) = \operatorname{top\text{-}k}\bigl\{c_{i,j} : \text{sim}(\phi(q'), \mathbf{v}_{i,j}) \bigr\}
\end{equation}
where $q'$ is a rewritten version of query $q$ and $\text{sim}(\cdot,\cdot)$ denotes a similarity function. The answer is then generated as:
\begin{equation}
    a = \text{LLM}\bigl(q, \mathcal{R}_{\text{CBR}}(q)\bigr)
\end{equation}

In our implementation, retrieval is further enhanced with full-text search, rank fusion, and neural reranking (detailed in Section~\ref{sec:cbr_arch}).

\subsubsection{Semantic File Routing (SFR)}

The SFR paradigm operates on whole documents and bypasses offline indexing entirely. At query time, an LLM with structured output extracts document-identifying metadata from the query:
\begin{equation}\label{eq:sfr_extraction}
    \text{LLM}_{\text{parse}}(q) \rightarrow \bigl\{(t_1, y_1), (t_2, y_2), \ldots, (t_K, y_K)\bigr\}
\end{equation}
where each $(t_k, y_k)$ is a (ticker, year) pair. These pairs are resolved to file paths via the mapping in Eq.~\eqref{eq:doc_mapping}:
\begin{equation}
    \mathcal{R}_{\text{SFR}}(q) = \bigl\{f(t_k, y_k) : k = 1, \ldots, K\bigr\}
\end{equation}

The answer is generated with the full document(s) as context:
\begin{equation}
    a = \text{LLM}\bigl(q, \mathcal{R}_{\text{SFR}}(q)\bigr)
\end{equation}

\subsubsection{Hybrid Document-Routed Retrieval (HDRR)}

The hybrid paradigm combines document-level routing with chunk-level retrieval. It uses SFR's metadata extraction (Eq.~\ref{eq:sfr_extraction}) to identify the target document(s), then restricts CBR's chunk-based retrieval (Eq.~\ref{eq:cbr_retrieval}) to only the chunks belonging to those documents:
\begin{equation}\label{eq:hdrr_retrieval}
    \mathcal{R}_{\text{HDRR}}(q) = \operatorname{top\text{-}k}\bigl\{c_{i,j} : d_i \in \mathcal{R}_{\text{SFR}}(q),\; \text{sim}(\phi(q'), \mathbf{v}_{i,j}) \bigr\}
\end{equation}

When document routing fails (no documents resolved), HDRR falls back to unrestricted chunk retrieval over the full corpus, equivalent to standard CBR:
\begin{equation}
    \mathcal{R}_{\text{HDRR}}^{\text{fallback}}(q) = \mathcal{R}_{\text{CBR}}(q)
\end{equation}

The answer is generated from the scoped chunk context:
\begin{equation}
    a = \text{LLM}\bigl(q, \mathcal{R}_{\text{HDRR}}(q)\bigr)
\end{equation}

HDRR inherits SFR's immunity to cross-document chunk confusion (by restricting retrieval to the correct document) while preserving CBR's ability to surface precisely relevant text fragments (by performing chunk-level retrieval within the scoped document).

\subsection{Corpus Structure Assumption}\label{sec:corpus_assumption}

SFR relies on two structural properties of the corpus:

\begin{definition}[Naming Regularity]\label{def:naming}
A document corpus $\mathcal{D}$ exhibits \emph{naming regularity} if there exists a deterministic mapping $g: \mathcal{M} \rightarrow \text{Paths}$ from a metadata space $\mathcal{M}$ to file-system paths, such that $g$ is injective and the metadata $m_i \in \mathcal{M}$ for each document $d_i$ can be reliably extracted from natural language queries about $d_i$.
\end{definition}

\begin{definition}[Structural Homogeneity]\label{def:homogeneity}
A document corpus $\mathcal{D}$ exhibits \emph{structural homogeneity} if the documents share a common organizational template, such that analogous sections across different documents contain semantically similar but entity-distinct content.
\end{definition}

Financial regulatory filings naturally satisfy both properties: 10-K reports follow SEC-mandated structures (naming regularity via ticker and year), and all reports contain standardized sections such as ``Risk Factors,'' ``MD\&A,'' and ``Financial Statements'' (structural homogeneity). Our hypothesis is that structural homogeneity creates retrieval confusion for chunk-based systems while naming regularity enables SFR to bypass this confusion entirely.

\subsection{Evaluation Framework}

We evaluate both paradigms using an LLM-as-judge approach~\citep{zheng2023llmjudge}, where a separate LLM instance scores the generated answer $a$ against a ground-truth reference answer $a^*$ on a scale of 1 to 10:
\begin{equation}
    s = \text{LLM}_{\text{eval}}(q, a, a^*) \in \{1, 2, \ldots, 10\}
\end{equation}

We report four complementary metrics:
\begin{itemize}
    \item \textbf{Average Score}: $\bar{s} = \frac{1}{|\mathcal{Q}|}\sum_{q \in \mathcal{Q}} s_q$, reflecting overall answer quality.
    \item \textbf{Failure Rate}: $P(s = 1)$, measuring the proportion of completely incorrect answers.
    \item \textbf{Correctness Rate}: $P(s \geq 8)$, measuring the proportion of substantively correct answers.
    \item \textbf{Perfect Rate}: $P(s = 10)$, measuring the proportion of fully correct and complete answers.
\end{itemize}

These metrics capture different aspects of system behavior: average score reflects central tendency, failure rate measures worst-case performance, correctness rate measures practical utility, and perfect rate measures ceiling performance.

\section{System Architectures}\label{sec:architectures}

This section describes the three RAG architectures compared in our study. All systems share the same generation model (GPT-4.1) and evaluation pipeline, differing only in how they retrieve relevant context for a given query. The first paradigm, Chunk-Based Retrieval (CBR), follows the canonical RAG design: documents are split into overlapping fragments, embedded into a shared vector space, and retrieved via hybrid search and neural reranking. The second paradigm, Semantic File Routing (SFR), bypasses chunking entirely by using LLM structured output to resolve document identity from the query and providing the whole document as context. The third paradigm, Hybrid Document-Routed Retrieval (HDRR), combines both: it uses SFR's routing step to identify the target document, then applies CBR's chunk-level retrieval scoped to that document. This progression---from unconstrained to document-scoped chunk retrieval---directly addresses the cross-document confusion problem identified in Section~\ref{sec:introduction}.

\subsection{Chunk-Based RAG (CBR)}\label{sec:cbr_arch}

\begin{figure}[H]
\centering
\includegraphics[width=0.9\textwidth]{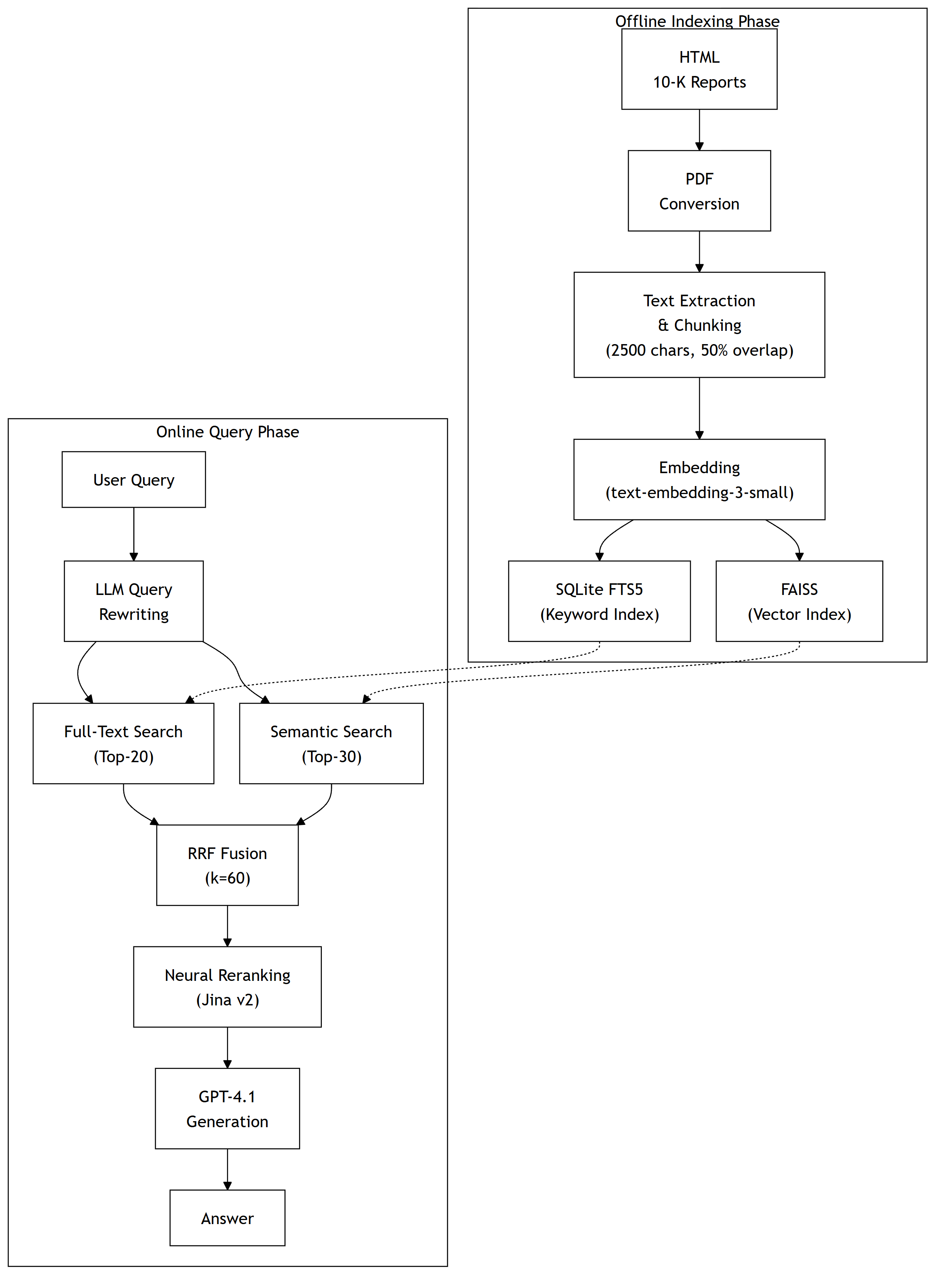}
\caption{Chunk-Based RAG (CBR) pipeline. The offline phase indexes documents into dual stores (FTS5 for keyword search, FAISS for semantic search). The online phase retrieves, fuses, reranks, and generates.}\label{fig:cbr_pipeline}
\end{figure}

The chunk-based system implements a full retrieval pipeline with offline indexing and online query processing, as shown in Fig.~\ref{fig:cbr_pipeline}. It serves as the primary baseline in our study and represents the dominant paradigm for RAG over large document collections~\citep{lewis2020rag, gao2023retrieval}. The pipeline is divided into two phases: an offline \emph{indexing} phase that preprocesses the entire corpus once, and an online \emph{query} phase that retrieves and generates answers at inference time.

\subsubsection{Offline Indexing Pipeline}

\textbf{Document Conversion.}
The 500 S\&P~500 10-K reports are obtained in HTML format from company filings and converted to PDF using Playwright~\citep{playwright2024}, a browser automation tool that renders HTML with proper formatting and layout preservation. This conversion ensures consistent text extraction across documents.

\textbf{Text Extraction and Chunking.}
PDF documents are parsed using PyMuPDF to extract textual content while preserving document structure. We employ fixed-size chunking with a chunk size of 2,500 characters and 50\% overlap (1,250 characters). Fixed-size chunking was selected over semantic or recursive strategies for reproducibility and computational efficiency across a 500-document corpus. The chunk size of 2,500 characters (approximately 500--600 tokens) is calibrated to fit comfortably within the input limit of the embedding model while retaining sufficient local context for meaningful semantic representation~\citep{transformerecommerce}. The 50\% overlap ensures that information spanning chunk boundaries---such as multi-paragraph tables, footnoted financial figures, or management commentary that continues across page breaks---is captured in at least one chunk without duplication artifacts.

\textbf{Dual-Index Storage.}
Extracted chunks are stored in two complementary index structures:
\begin{itemize}
    \item \emph{SQLite FTS5}~\citep{sqlite2024fts5}: A full-text search index supporting Boolean queries, phrase matching, and BM25-based relevance ranking for keyword retrieval.
    \item \emph{FAISS}~\citep{johnson2021faiss}: A vector index storing 1,024-dimensional embeddings generated by OpenAI's text-embedding-3-small model~\citep{openai2024embeddings}, enabling nearest neighbor search for semantic retrieval.
\end{itemize}
The FAISS index uses an exact inner-product search over L2-normalized vectors (IndexFlatIP), trading computational efficiency for retrieval precision on a corpus of manageable scale ($\sim$500K chunks). This avoids the quantization errors introduced by approximate methods such as IVF or HNSW, which are better suited for billion-scale corpora. Together, the FTS5 and FAISS indices provide complementary retrieval coverage: BM25-based full-text search excels at precise entity matching---ticker symbols, monetary figures, and SEC section identifiers---while semantic search captures paraphrase and intent variations that keyword lookup cannot handle.

\subsubsection{Online Query Pipeline}

\textbf{Query Rewriting.}
Raw user queries are processed by GPT-4.1 to produce two outputs: (1)~a clarified query for semantic search and (2)~a list of extracted keywords for keyword-based retrieval. This preprocessing step addresses query ambiguity, corrects grammatical issues, and improves retrieval recall.

\textbf{Hybrid Retrieval.}
Two parallel retrieval paths are executed:
\begin{itemize}
    \item \emph{Full-Text Search (FTS):} The extracted keywords are used to query the SQLite FTS5 index, returning the top-20 most relevant chunks.
    \item \emph{Semantic Search:} The clarified query is embedded and used for approximate nearest neighbor search in the FAISS index, returning the top-30 semantically similar chunks with a distance threshold of 2.0.
\end{itemize}

\textbf{Reciprocal Rank Fusion.}
Results from both retrieval methods are combined using RRF~\citep{cormack2009rrf} with $k = 60$:
\begin{equation}
    \text{RRFscore}(d) = \frac{1}{60 + r_{\text{FTS}}(d)} + \frac{1}{60 + r_{\text{sem}}(d)}
\end{equation}
where $r_{\text{FTS}}(d)$ and $r_{\text{sem}}(d)$ are the ranks of chunk $d$ in the FTS and semantic result lists, respectively.

\textbf{Neural Reranking.}
The fused candidate list (up to 30 chunks) is re-scored by Jina Reranker v2~\citep{jina2024reranker}, a 278M-parameter cross-encoder model. Two adaptive cutoff strategies select the final context:
\begin{itemize}
    \item \emph{Cumulative Probability Threshold:} Chunks are retained until the cumulative normalized score reaches 55\%.
    \item \emph{Score Cliff Detection:} If a chunk's score drops by more than 0.15 from the top score, it and all lower-ranked chunks are excluded.
\end{itemize}
This typically yields 5--10 high-quality chunks for generation.

\textbf{Answer Generation.}
The selected chunks are concatenated into a context string and provided to GPT-4.1 with a system prompt instructing the model to answer based solely on the provided context.

\subsubsection{Design Rationale and Limitations}

The CBR pipeline reflects established best practices for dense retrieval over large document collections~\citep{karpukhin2020dense, gao2023retrieval}. The dual-index architecture (FTS5 + FAISS) captures complementary retrieval signals: keyword matching excels at entity names and financial identifiers such as ticker symbols, while semantic search handles paraphrase and intent variations. RRF fusion is unsupervised and rank-based, making it robust to score scale differences between the two retrievers~\citep{cormack2009rrf}. Adaptive cutoff reranking ensures that the generation model receives a compact, high-precision context window rather than a fixed-size candidate set.

However, CBR's core assumption---that relevant chunks can be identified from a global embedding space shared by all documents---breaks down in structurally homogeneous corpora. When hundreds of 10-K filings are indexed together, semantically near-identical sections from different companies occupy adjacent regions of the embedding space. This cross-document chunk confusion is the primary failure mode analyzed in Section~\ref{sec:discussion} and the principal motivation for the routing-based architectures introduced below.

\clearpage
\onecolumn
\twocolumn

\subsection{Semantic File Routing (SFR)}\label{sec:sfr_arch}

The SFR system eliminates the offline indexing phase entirely. It requires only that documents be organized in a directory structure following a predictable naming convention. Fig.~\ref{fig:sfr_pipeline} illustrates the pipeline.

\medskip
\noindent\begin{minipage}{\columnwidth}
\centering
\includegraphics[width=1.0\columnwidth]{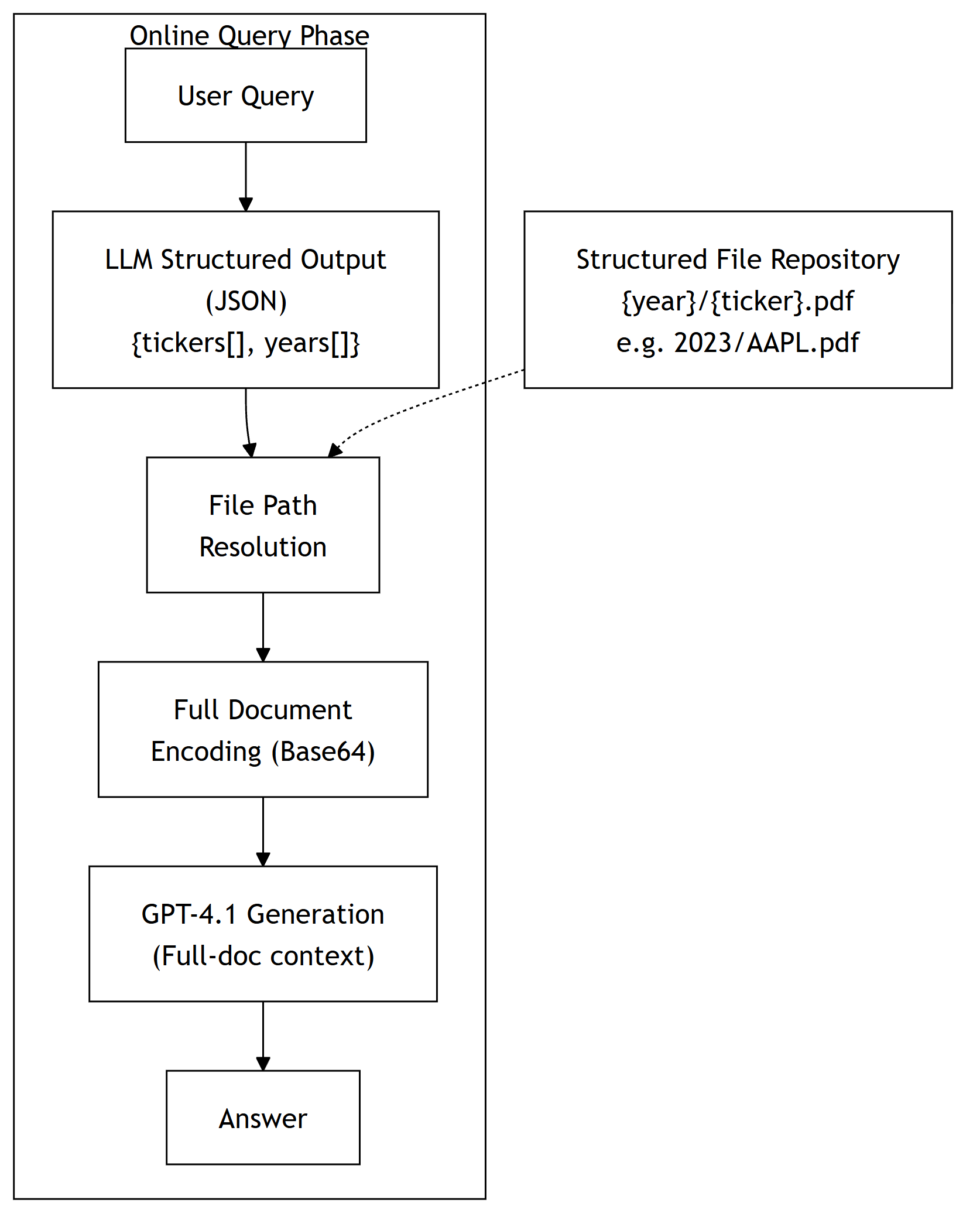}
\figcaption{Semantic File Routing (SFR) pipeline. No offline indexing is required. The query is parsed into structured metadata, resolved to file paths, and the full document is provided to the generation model.}\label{fig:sfr_pipeline}
\end{minipage}
\medskip

\subsubsection{Corpus Organization}

Documents are stored in a hierarchical directory following the pattern:
\begin{center}
\texttt{data/\{year\}/\{ticker\}.\{ext\}}
\end{center}
where \texttt{year} is the fiscal year, \texttt{ticker} is the company's stock ticker symbol, and \texttt{ext} is the file extension (\texttt{pdf} or \texttt{txt}). For the S\&P~500 corpus, this results in a flat hierarchy with approximately 500 files per year.

\subsubsection{Online Query Pipeline}

\textbf{Metadata Extraction via Structured Output.}
The query $q$ is sent to GPT-4.1 configured with JSON structured output mode~\citep{openai2024structured}. The model is instructed to extract:
\begin{itemize}
    \item \texttt{tickers}: an array of company ticker symbols referenced in the query.
    \item \texttt{years}: an array of fiscal years referenced in the query, or a default year if none is specified.
\end{itemize}

The structured output guarantee ensures that the response always conforms to the expected JSON schema, eliminating parsing failures. An example extraction is shown in Algorithm~\ref{alg:sfr_extraction}.

\begin{algorithm}[t]
\caption{Semantic File Routing: Query Processing}\label{alg:sfr_extraction}
\begin{algorithmic}[1]
\Require Query $q$, data directory $D$, default year $y_0$
\Ensure Answer $a$
\State $\textit{metadata} \gets \text{LLM}_{\text{parse}}(q)$ \Comment{JSON structured output}
\State $\textit{tickers} \gets \textit{metadata}.\text{tickers}$
\State $\textit{years} \gets \textit{metadata}.\text{years}$
\If{$\textit{years} = \emptyset$}
    \State $\textit{years} \gets [y_0]$
\EndIf
\State $\textit{files} \gets \emptyset$
\For{each $(t, y)$ in $\textit{tickers} \times \textit{years}$}
    \State $p \gets D / y / (t + \texttt{".pdf"})$
    \If{$\text{exists}(p)$}
        \State $\textit{files} \gets \textit{files} \cup \{p\}$
    \Else
        \State $p \gets D / y / (t + \texttt{".txt"})$
        \If{$\text{exists}(p)$} $\textit{files} \gets \textit{files} \cup \{p\}$
        \EndIf
    \EndIf
\EndFor
\If{$\textit{files} = \emptyset$ \textbf{and} years $\neq [y_0]$}
    \State Retry with $\textit{years} \gets [y_0]$ \Comment{Fallback}
\EndIf
\State $\textit{context} \gets \text{encode\_files}(\textit{files})$ \Comment{Base64 encoding}
\State $a \gets \text{LLM}_{\text{gen}}(q, \textit{context})$
\State \Return $a$
\end{algorithmic}
\end{algorithm}

\textbf{File Resolution and Fallback.}
For each extracted (ticker, year) pair, the system constructs a file path and checks for existence, trying PDF first and falling back to TXT. If no files are found for the extracted years, the system retries with a configurable default year (2023 in our experiments). This fallback mechanism handles queries that do not specify a fiscal year explicitly.

\textbf{Document Encoding.}
Located files are encoded in base64 format and attached to the API request as file inputs, leveraging the multimodal file-reading capabilities of modern LLM APIs. This enables the model to process PDF documents directly, including formatted text, tables, and layout information that would be lost in text-only extraction.

\textbf{Answer Generation.}
The generation model (GPT-4.1) receives the original query along with the full document file(s) and produces an answer. Unlike the CBR system where context is limited to selected chunks, SFR provides the complete document, allowing the model to reason over the full scope of the filing.

\subsection{Hybrid Document-Routed Retrieval (HDRR)}\label{sec:hdrr_arch}

\begin{figure*}[!t]
\centering
\includegraphics[width=0.8\textwidth]{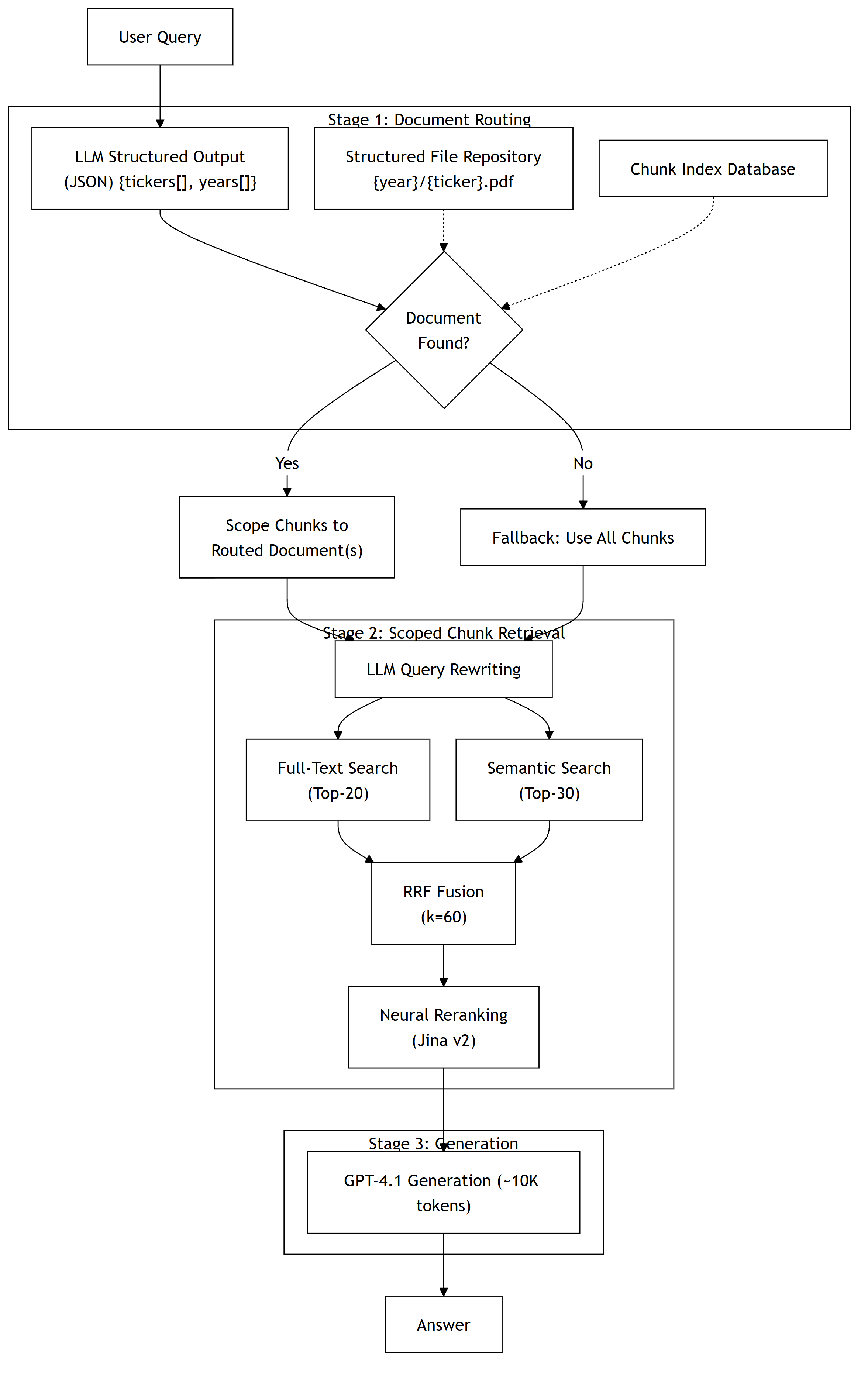}
\caption{Hybrid Document-Routed Retrieval (HDRR) pipeline. Stage~1 routes the query to the correct document via LLM structured output. Stage~2 performs scoped hybrid search (FTS + semantic + RRF + reranking) restricted to the routed document's chunks. If routing fails, the system falls back to full-corpus retrieval.}\label{fig:hdrr_pipeline}
\end{figure*}

The HDRR system combines SFR's document routing with CBR's chunk-level retrieval. It reuses the same offline index as CBR (no additional indexing required) and adds a document-routing stage before retrieval. Fig.~\ref{fig:hdrr_pipeline} illustrates the pipeline and Algorithm~\ref{alg:hdrr} provides the pseudocode.

\begin{algorithm}[t]
\caption{Hybrid Document-Routed Retrieval: Query Processing}\label{alg:hdrr}
\begin{algorithmic}[1]
\Require Query $q$, data directory $D$, default year $y_0$, chunk index $\mathcal{I}$
\Ensure Answer $a$
\State \textit{metadata} $\gets \text{LLM}_{\text{parse}}(q)$ \Comment{Stage 1: Document Routing}
\State \textit{tickers} $\gets$ \textit{metadata}.\text{tickers}
\State \textit{years} $\gets$ \textit{metadata}.\text{years}
\If{\textit{years} $= \emptyset$}
    \State \textit{years} $\gets [y_0]$
\EndIf
\State $\textit{doc\_ids} \gets \emptyset$
\For{each $(t, y)$ in \textit{tickers} $\times$ \textit{years}}
    \State $d \gets \text{lookup\_document}(D, y, t)$
    \If{$d \neq \text{null}$}
        \State $\textit{doc\_ids} \gets \textit{doc\_ids} \cup \{\text{id}(d)\}$
    \EndIf
\EndFor
\If{$\textit{doc\_ids} = \emptyset$}
    \State $\textit{scope} \gets \mathcal{I}$ \Comment{Fallback: full-corpus search}
\Else
    \State $\textit{scope} \gets \{c \in \mathcal{I} : \text{doc}(c) \in \textit{doc\_ids}\}$ \Comment{Scoped index}
\EndIf
\State $q' \gets \text{LLM}_{\text{rewrite}}(q)$ \Comment{Stage 2: Scoped Chunk Retrieval}
\State $R_{\text{fts}} \gets \text{FTS}(q', \textit{scope})$
\State $R_{\text{sem}} \gets \text{SemSearch}(q', \textit{scope})$
\State $R_{\text{fused}} \gets \text{RRF}(R_{\text{fts}}, R_{\text{sem}})$
\State $R_{\text{final}} \gets \text{Rerank}(q', R_{\text{fused}})$
\State $a \gets \text{LLM}_{\text{gen}}(q, R_{\text{final}})$ \Comment{Stage 3: Answer Generation}
\State \Return $a$
\end{algorithmic}
\end{algorithm}

\subsubsection{Stage 1: Document Routing}

Identical to SFR, the query is processed by GPT-4.1 with structured output to extract ticker symbols and fiscal years. The extracted metadata is resolved to document IDs in the existing chunk index database. If exact-path lookup fails, the system falls back to filename-based matching. If no documents can be identified at all, the system degrades gracefully to full-corpus retrieval (equivalent to standard CBR), ensuring that HDRR never performs worse than CBR due to routing failures.

\subsubsection{Stage 2: Scoped Chunk Retrieval}

The full CBR retrieval pipeline, comprising query rewriting, hybrid search (FTS + semantic), RRF fusion, and neural reranking, is executed with one critical modification: all retrieval operations are restricted to chunks belonging to the routed document(s). This scoping is implemented by filtering the chunk ID space before both FTS and FAISS queries, requiring no changes to the underlying search algorithms.

By restricting the search space to a single document (typically 50--200 chunks), HDRR eliminates cross-document chunk confusion by construction. The reranker operates on candidates that all originate from the correct document, enabling it to focus on selecting the most query-relevant passages rather than filtering out cross-document noise.

\subsubsection{Stage 3: Answer Generation}

The reranked chunks are provided to GPT-4.1 with the same prompt template as CBR. Because the chunks are drawn from the correct document and selected by the reranker for relevance, the generation model receives compact, precisely targeted context that combines SFR's document-level accuracy with CBR's chunk-level precision.

\subsection{Architectural Comparison}

Table~\ref{tab:arch_comparison} summarizes the key architectural differences between the three paradigms.

\begin{table*}[t]
\caption{Architectural comparison of CBR, SFR, and HDRR paradigms.}\label{tab:arch_comparison}
\centering
\small
\begin{tabular*}{\tblwidth}{@{}LLLL@{}}
\toprule
\textbf{Property} & \textbf{CBR} & \textbf{SFR} & \textbf{HDRR} \\
\midrule
Offline indexing   & Required & Not required & Required (shared with CBR) \\
Retrieval unit     & Chunk & Whole document & Document $\to$ Chunk \\
Retrieval method   & Hybrid + rerank & LLM structured output & Routing + scoped hybrid \\
Context scope      & 5--10 chunks & Full document(s) & 5--10 scoped chunks \\
Vector database    & Required & Not required & Required (shared with CBR) \\
Corpus assumption  & None & Naming regularity & Naming regularity \\
Cross-doc confusion & Possible & Eliminated & Eliminated \\
Context dilution   & None & Significant & None \\
Generation tokens  & $\sim$5K--15K & $\sim$50K--200K & $\sim$5K--15K \\
\bottomrule
\end{tabular*}
\end{table*}

The key insight is that HDRR inherits the strengths of both paradigms while avoiding their weaknesses: it eliminates cross-document confusion (like SFR) and avoids context dilution (like CBR). The cost is an additional LLM API call for document routing and the requirement that the corpus exhibit naming regularity.

\section{Experimental Setup}\label{sec:experiments}

\subsection{Dataset}

We evaluate all three systems using the FinDER benchmark~\citep{finder2025}, a dataset specifically designed for financial question answering and RAG evaluation.

\textbf{Dataset Characteristics.}
FinDER contains 5,703 query-evidence-answer triplets derived from real-world financial inquiries across 10-K filings of S\&P~500 companies. The dataset exhibits several properties that make it a challenging and realistic benchmark:
\begin{itemize}
    \item \textbf{Question type distribution:} 84.5\% qualitative questions (requiring narrative or explanatory answers) and 15.5\% quantitative questions (requiring numerical answers, with half involving multi-step reasoning).
    \item \textbf{Query characteristics:} Concise, analyst-style queries with domain-specific abbreviations, jargon, and frequently ambiguous references.
    \item \textbf{Evidence complexity:} Relevant evidence is often scattered across multiple non-adjacent sections within a filing, requiring synthesis and cross-referencing.
\end{itemize}

\subsection{Sampling Strategy}

Following the methodology of our prior work~\citep{cheng2026reranking}, we employ stratified random sampling to generate five independent test groups. Each group contains 300 queries, sampled uniformly at random (approximately 5\%) from the full FinDER dataset of 5,703 queries. This yields a total evaluation set of 1,500 queries.

Critically, the same five sample groups are used for \emph{every} system evaluation reported in this paper---the three original paradigms (CBR, SFR, HDRR) and the three additional comparison systems introduced in this revision (CBR+Meta, CBR+LLM-Ctx, Agentic). This paired experimental design ensures that observed performance differences are attributable to the retrieval paradigm rather than sampling variation. The multi-group design also enables assessment of inter-group consistency and estimation of result variance. Per-system aggregates are computed over the full 1,500-query set unless otherwise noted; the cross-evaluator subset described in Section~\ref{sec:eval_robustness} is a stratified subsample of these same groups.

\subsection{System Configurations}

\textbf{Chunk-Based RAG (CBR).}
The CBR system is configured identically to the ``with reranking'' configuration from our prior study~\citep{cheng2026reranking}, representing the strongest variant of the chunk-based pipeline. The complete configuration is provided in Appendix~\ref{app:hyperparams}, Table~\ref{tab:cbr_params}.

\textbf{Semantic File Routing (SFR).}
The SFR system uses GPT-4.1 for both query parsing (structured output) and answer generation. The file repository contains 10-K reports organized as \texttt{\{year\}/\{ticker\}.pdf}, with a default year of 2023. The complete configuration is provided in Appendix~\ref{app:hyperparams}, Table~\ref{tab:sfr_params}.

\textbf{Hybrid Document-Routed Retrieval (HDRR).}
The HDRR system combines SFR's document routing (Stage~1) with CBR's full retrieval pipeline (Stage~2). It uses the same chunk index as CBR (no additional indexing) and the same structured output parsing as SFR. When document routing fails, HDRR falls back to full-corpus chunk retrieval. The complete configuration is provided in Appendix~\ref{app:hyperparams}, Table~\ref{tab:hdrr_params}.

\textbf{CBR+Meta (V-A): metadata-prefix contextual.}
V-A is the cheapest form of contextual indexing: for every chunk whose parent document is at \texttt{\{year\}/\{ticker\}.pdf}, we prepend the static string ``\texttt{[Company: \{TICKER\}, Fiscal Year: \{YEAR\}]}'' to the chunk text before computing its embedding and inserting it into the FTS index. The remainder of the pipeline (RRF fusion, neural reranking, chunk-conditioned generation) is identical to CBR. This isolates the contribution of pure document-identity tagging---zero LLM cost at indexing, deterministic, easy to reproduce---and serves as the lower-bound contextual baseline.

\textbf{CBR+LLM-Ctx (V-B): LLM-generated contextual~\citep{anthropic2024contextual}.}
V-B is the faithful Contextual Retrieval baseline. For each chunk, a 50--100 token paragraph situating that chunk inside its parent document (company, fiscal year, section, topic) is generated by GPT-4.1-mini at temperature 0.0 and prepended to the chunk text, separated by ``\texttt{\textbackslash n---\textbackslash n}''. We use OpenAI prompt caching on the full document text to amortize input cost across the document's chunks. The generated context is persisted in SQLite alongside the chunk so that the index is auditable. As with V-A, the rest of the pipeline matches CBR. One-time indexing cost on our 497-document corpus (169{,}640 chunks) was bounded by a \$100 budget cap and came in well under it; per-query cost matches CBR.

\textbf{Agentic RAG.}
The Agentic system replaces HDRR's upfront routing with a post-hoc verification loop, instantiating the Self-RAG-style pattern~\citep{asai2024selfrag} described in Section~\ref{sec:rw_agentic}. For each query: (i)~a ticker is extracted from the query using the same structured-output prompt as SFR; (ii)~full-corpus CBR retrieves an initial chunk set; (iii)~a single LLM call verifies whether the first three reranked chunks ``primarily contain information about'' the named ticker, returning \texttt{\{"belongs": bool, "reason": str\}}; (iv)~on \texttt{false}, the query is re-anchored with the prefix ``In \texttt{\{ticker\}}'s 10-K annual report: '' and CBR is re-run, with a retry cap of 2. If no ticker is extracted (mirroring HDRR's fallback condition), verification is skipped and the system reduces to plain CBR. This design isolates the cost and accuracy of post-hoc verification against the upfront-routing approach of HDRR.

All systems use GPT-4.1 as the generation model with temperature set to 0.0 for deterministic output, ensuring that differences in answer quality are attributable to the retrieval mechanism rather than generation stochasticity. Document routing (where used) also uses GPT-4.1 at temperature 0.0; the V-B contextual generator uses GPT-4.1-mini at temperature 0.0.

\subsection{Evaluation Methodology}

\subsubsection{LLM-as-Judge Scoring}

We employ an LLM-as-judge approach~\citep{zheng2023llmjudge, li2025generation} using GPT-4.1 as the evaluation model. For each query, the evaluator receives three inputs: (1)~the original query, (2)~the ground-truth answer from FinDER, and (3)~the system-generated answer. The evaluator assigns a score on a 1--10 integer scale using the following rubric:

\begin{itemize}
    \item \textbf{Score 1:} Completely incorrect or unrelated answer.
    \item \textbf{Score 5:} Partially correct with significant missing information.
    \item \textbf{Score 8:} Substantively correct with minor omissions or imprecisions.
    \item \textbf{Score 10:} Fully correct and complete answer.
\end{itemize}

The evaluator is configured with temperature 0.0 and produces structured JSON output containing the score and a brief justification. This ensures deterministic, reproducible evaluation.

\subsubsection{Evaluation Metrics}

We report four metrics designed to capture different aspects of system behavior:

\begin{enumerate}
    \item \textbf{Average Score} ($\bar{s}$): The arithmetic mean of all evaluation scores within a group, reflecting overall answer quality on the 1--10 scale.
    \item \textbf{Failure Rate} ($P(s = 1)$): The percentage of queries receiving the minimum score, indicating completely incorrect answers. This metric measures worst-case behavior and system reliability.
    \item \textbf{Correctness Rate} ($P(s \geq 8)$): The percentage of queries receiving a score of 8 or above, indicating substantively correct answers. This metric reflects practical utility, measured as the proportion of queries where the system provides a usable answer.
    \item \textbf{Perfect Rate} ($P(s = 10)$): The percentage of queries receiving the maximum score, indicating fully correct and complete answers. This metric measures ceiling performance.
\end{enumerate}

\subsubsection{Statistical Controls}

The five-group experimental design serves multiple purposes:
\begin{itemize}
    \item \textbf{Variance estimation:} Inter-group variance provides a measure of result stability under different query samples.
    \item \textbf{Paired comparison:} Using the same groups for both systems enables paired statistical analysis, reducing the influence of query difficulty on comparative conclusions.
    \item \textbf{Reproducibility:} Multiple independent groups demonstrate that findings are not artifacts of a particular query selection.
\end{itemize}

\subsubsection{LLM-as-Judge Symmetry and Cross-Evaluator Robustness}\label{sec:eval_robustness}

\textbf{Symmetry argument.}
The same GPT-4.1 instance is used for generation, routing (in SFR/HDRR/Agentic), and scoring. This introduces a potential circular-evaluation risk: any systematic stylistic preference the evaluator holds could in principle inflate the apparent quality of answers produced by the same model family. The mitigating structural property is \emph{symmetry}: GPT-4.1 produces and scores every system's answers under identical evaluator configuration, so any systematic stylistic preference applies uniformly across CBR, SFR, HDRR, V-A, V-B, and Agentic. Our claims are \emph{comparative} (HDRR $>$ alternatives on the four metrics) rather than absolute (HDRR achieves quality score $X$); a uniform stylistic offset therefore cancels in the comparison on which the paper's conclusions depend.

\textbf{Empirical cross-evaluator check.}
Symmetry alone is an argument, not evidence. To produce direct evidence we re-scored a stratified subset using a different model family. The procedure: draw 20 queries per group from the five 300-query splits with a fixed seed; one query was drawn into two splits, yielding 99 unique IDs; re-score every system's answers for those 99 queries using Anthropic's \texttt{claude-sonnet-4-6}~\citep{anthropic2024claude} with the \emph{identical} 1--10 rubric prompt used by GPT-4.1 (no rubric drift). This produces 6~systems $\times$ 99~queries $=$ 594 cross-evaluator scores. The Anthropic client is wired through a thin facade that mirrors the OpenAI evaluator's interface and floors the response budget to accommodate Claude's preamble-before-JSON behavior; all 594 calls returned valid scores.

For each system we compute (i)~per-query Spearman rank correlation $\rho$ between the GPT-4.1 score and the Claude score; (ii)~mean absolute score difference; (iii)~per-evaluator failure (s$=$1) and correctness (s$\geq$7) rates. At the system level, we additionally compute the Spearman rank correlation between the two evaluators over the six per-system average scores, and check whether the system-level ranking induced by GPT-4.1 is preserved under Claude. Results appear in Section~\ref{sec:cross_evaluator}.

\FloatBarrier

\section{Results}\label{sec:results}

\subsection{Overall Performance Comparison}

Table~\ref{tab:main_results} presents the complete results for the three core paradigms (CBR, SFR, HDRR) across the five experimental groups; the three additional comparison systems requested by reviewers (V-A, V-B, Agentic) are reported alongside in Section~\ref{sec:extended_baselines}, Table~\ref{tab:extended_results}. Fig.~\ref{fig:comparison} provides a side-by-side overview of the CBR and SFR paradigms whose trade-off motivates the hybrid approach.

\noindent\begin{minipage}{\columnwidth}
\centering
\includegraphics[width=\columnwidth]{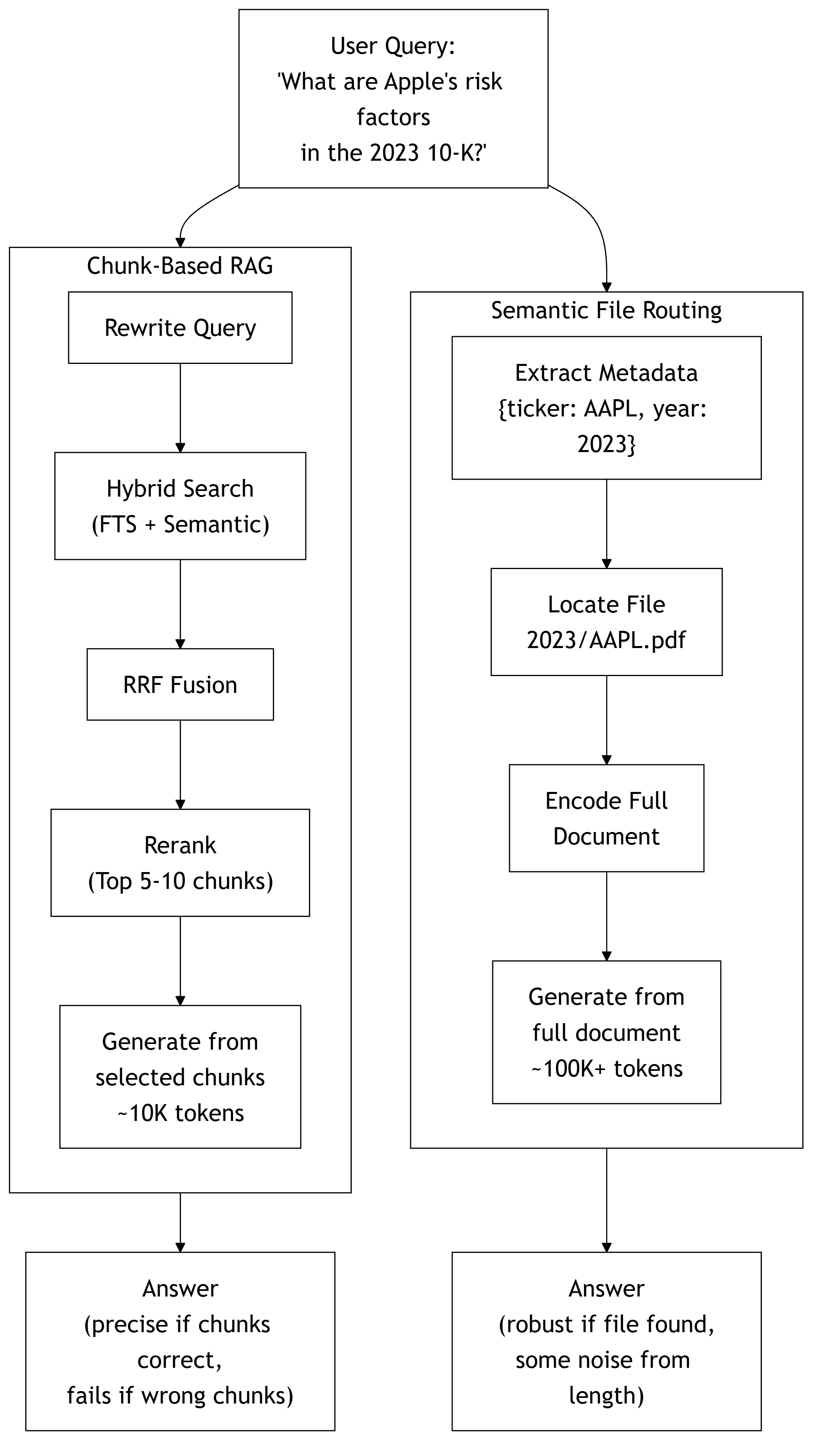}
\figcaption{Side-by-side comparison of the CBR and SFR paradigms for the same query. CBR retrieves targeted chunks ($\sim$10K tokens) while SFR provides the full document ($\sim$100K+ tokens), illustrating the precision-vs-robustness trade-off that HDRR resolves.}\label{fig:comparison}
\end{minipage}

\begin{table*}[t]
\caption{Performance comparison across five experimental groups. CBR = Chunk-Based RAG; SFR = Semantic File Routing; HDRR = Hybrid Document-Routed Retrieval. Best values per metric are \textbf{bolded}.}\label{tab:main_results}
\centering
\small
\begin{tabular*}{\tblwidth}{@{}CCLLLL@{}}
\toprule
\textbf{Group} & \textbf{System} & \textbf{Avg Score} & \textbf{$s{=}1$ (\%)} & \textbf{$s{\geq}8$ (\%)} & \textbf{$s{=}10$ (\%)} \\
\midrule
\multirow{3}{*}{01} & CBR  & 5.85 & 24.3 & 47.7 & 12.0 \\
                     & SFR  & 6.53 & 9.3  & 44.1 & 7.7  \\
                     & HDRR & \textbf{7.44} & \textbf{7.0} & \textbf{64.7} & \textbf{20.0} \\
\midrule
\multirow{3}{*}{02} & CBR  & 6.17 & 21.7 & 51.6 & 16.0 \\
                     & SFR  & 6.61 & 9.3  & 46.4 & 8.7  \\
                     & HDRR & \textbf{7.76} & \textbf{5.0} & \textbf{72.7} & \textbf{19.0} \\
\midrule
\multirow{3}{*}{03} & CBR  & 5.64 & 25.7 & 44.0 & 14.0 \\
                     & SFR  & 6.11 & 12.0 & 36.7 & 10.7 \\
                     & HDRR & \textbf{7.39} & \textbf{7.3} & \textbf{63.0} & \textbf{21.7} \\
\midrule
\multirow{3}{*}{04} & CBR  & 6.21 & 21.0 & 51.3 & 14.7 \\
                     & SFR  & 6.67 & 9.7  & 50.6 & 10.0 \\
                     & HDRR & \textbf{7.72} & \textbf{5.3} & \textbf{71.0} & \textbf{21.3} \\
\midrule
\multirow{3}{*}{05} & CBR  & 6.21 & 20.0 & 50.6 & 12.3 \\
                     & SFR  & 6.32 & 11.0 & 42.3 & 5.3  \\
                     & HDRR & \textbf{7.39} & \textbf{7.3} & \textbf{67.3} & \textbf{18.3} \\
\midrule
\multirow{3}{*}{\textbf{Mean}} & \textbf{CBR}  & 6.02 & 22.5 & 49.0 & 13.8 \\
                                & \textbf{SFR}  & 6.45 & 10.3 & 44.0 & 8.5  \\
                                & \textbf{HDRR} & \textbf{7.54} & \textbf{6.4} & \textbf{67.7} & \textbf{20.1} \\
\bottomrule
\end{tabular*}
\end{table*}

\subsection{Key Findings}

The results reveal a clear performance hierarchy, with HDRR dominating both baseline paradigms on every metric.

\textbf{Finding 1: The CBR--SFR robustness-precision trade-off.}
SFR achieves higher average scores (6.45 vs.\ 6.02) and far fewer catastrophic failures ($s{=}1$: 10.3\% vs.\ 22.5\%), but CBR produces more high-quality answers ($s{\geq}8$: 49.0\% vs.\ 44.0\%) and more perfect answers ($s{=}10$: 13.8\% vs.\ 8.5\%). Neither paradigm dominates the other across all metrics; root causes are analyzed in Section~\ref{sec:discussion}.

\textbf{Finding 2: HDRR resolves the trade-off.}
HDRR achieves the best performance on \emph{all four metrics} simultaneously. It attains the highest average score (7.54), the lowest failure rate (6.4\%), the highest correctness rate (67.7\%), and the highest perfect rate (20.1\%). This result holds in every individual group, not just on average.

\textbf{Finding 3: HDRR's improvement over CBR is substantial.}
Compared to CBR, HDRR improves average score by 1.52 points ($+$25.2\%), reduces failures by 16.1 percentage points ($-$71.6\% relative), increases correctness rate by 18.7~pp ($+$38.2\%), and increases perfect rate by 6.3~pp ($+$45.7\%).

\textbf{Finding 4: HDRR's improvement over SFR is equally substantial.}
Compared to SFR, HDRR improves average score by 1.09 points ($+$16.9\%), reduces failures by 3.9~pp ($-$37.9\% relative), increases correctness rate by 23.7~pp ($+$53.9\%), and increases perfect rate by 11.6~pp ($+$136.5\%).

\textbf{Finding 5: All three systems show consistent inter-group patterns.}
The standard deviation of average scores across the five groups is 0.24 for CBR, 0.22 for SFR, and 0.18 for HDRR. HDRR shows the lowest variance, indicating the most stable performance. The performance ranking (HDRR $>$ SFR $>$ CBR on average score; HDRR $>$ CBR $>$ SFR on perfect rate) is preserved across all five groups.

\subsection{Score Distribution Analysis}

Table~\ref{tab:distribution_shift} summarizes the distributional shift across all six systems using three score bands.

\begin{table}[t]
\caption{Score distribution across three bands---\textbf{Low} ($s \leq 3$), \textbf{Medium} ($4 \leq s \leq 7$), \textbf{High} ($s \geq 8$)---pooled percentages across the 1{,}500-query benchmark. Best values per column \textbf{bolded}.}\label{tab:distribution_shift}
\centering
\small
\setlength{\tabcolsep}{4pt}
\begin{tabular*}{\columnwidth}{@{\extracolsep{\fill}}lccc@{}}
\toprule
\textbf{System} & \textbf{Low} & \textbf{Medium} & \textbf{High} \\
\midrule
CBR               & 30.5 & 20.4 & 49.1 \\
SFR               & 16.7 & 39.3 & 44.0 \\
Agentic           & 25.5 & 23.1 & 51.5 \\
CBR+Meta (V-A)    & 19.0 & 23.1 & 57.9 \\
CBR+LLM-Ctx (V-B) & 15.9 & \textbf{19.5} & 64.5 \\
\textbf{HDRR}     & \textbf{12.3} & 19.9 & \textbf{67.7} \\
\bottomrule
\end{tabular*}
\end{table}

The distributional pattern reveals the mechanism behind each system's performance profile. SFR shifts probability mass from the low band to the medium band (reducing failures but not achieving high precision); its medium band swells by 18.9~pp over CBR while the high band moves only $-$5.1~pp. The contextual-indexing variants (V-A, V-B) shift mass away from the low band into the high band more directly than SFR but retain a non-trivial low band ($\geq 15.9$\%), because contextual prefixing improves discrimination but does not preclude wrong-document chunks from entering the candidate pool. Agentic's distribution most closely resembles CBR's, consistent with its post-hoc-verification design: most queries pass verification on the first attempt and inherit the underlying CBR distribution; retries recover only a subset of the low band. HDRR shifts mass directly from \emph{both} the low and medium bands into the high band, reducing the low band by 18.2~pp and channeling 18.7~pp into the high band over CBR. This pattern demonstrates that HDRR avoids the ``context dilution'' problem that causes SFR's mass to accumulate in the medium range and the ``residual wrong-document chunks'' problem that limits the contextual baselines.

\subsection{Paired Group Analysis}

Table~\ref{tab:paired_hdrr_cbr} shows the per-group performance differences between HDRR and CBR.

\begin{table}[t]
\caption{Per-group performance differences (HDRR $-$ CBR). Positive values on Avg Score and $s{\geq}8$/$s{=}10$ favor HDRR; negative values on $s{=}1$ favor HDRR.}\label{tab:paired_hdrr_cbr}
\centering
\small
\begin{tabular*}{\tblwidth}{@{}CCCCC@{}}
\toprule
\textbf{Group} & $\Delta$\textbf{Avg} & $\Delta$\textbf{$s{=}1$} & $\Delta$\textbf{$s{\geq}8$} & $\Delta$\textbf{$s{=}10$} \\
\midrule
01 & $+$1.59 & $-$17.3 & $+$17.0 & $+$8.0 \\
02 & $+$1.59 & $-$16.7 & $+$21.1 & $+$3.0 \\
03 & $+$1.75 & $-$18.4 & $+$19.0 & $+$7.7 \\
04 & $+$1.51 & $-$15.7 & $+$19.7 & $+$6.6 \\
05 & $+$1.18 & $-$12.7 & $+$16.7 & $+$6.0 \\
\midrule
\textbf{Mean} & $+$1.52 & $-$16.2 & $+$18.7 & $+$6.3 \\
\textbf{Std}  & 0.21   & 2.2    & 1.9    & 2.0 \\
\bottomrule
\end{tabular*}
\end{table}

HDRR improves over CBR on every metric in every group. The average score improvement ranges from $+$1.18 (Group~05) to $+$1.75 (Group~03), with low standard deviation (0.21), indicating a highly consistent improvement. The failure rate reduction is also consistent (mean $-$16.2~pp, std 2.2~pp), as is the correctness rate improvement (mean $+$18.7~pp, std 1.9~pp).

Table~\ref{tab:paired_hdrr_sfr} shows the per-group differences between HDRR and SFR.

\begin{table}[t]
\caption{Per-group performance differences (HDRR $-$ SFR). Positive values on Avg Score and $s{\geq}8$/$s{=}10$ favor HDRR; negative values on $s{=}1$ favor HDRR.}\label{tab:paired_hdrr_sfr}
\centering
\small
\begin{tabular*}{\tblwidth}{@{}CCCCC@{}}
\toprule
\textbf{Group} & $\Delta$\textbf{Avg} & $\Delta$\textbf{$s{=}1$} & $\Delta$\textbf{$s{\geq}8$} & $\Delta$\textbf{$s{=}10$} \\
\midrule
01 & $+$0.91 & $-$2.3 & $+$20.6 & $+$12.3 \\
02 & $+$1.15 & $-$4.3 & $+$26.3 & $+$10.3 \\
03 & $+$1.28 & $-$4.7 & $+$26.3 & $+$11.0 \\
04 & $+$1.05 & $-$4.4 & $+$20.4 & $+$11.3 \\
05 & $+$1.07 & $-$3.7 & $+$25.0 & $+$13.0 \\
\midrule
\textbf{Mean} & $+$1.09 & $-$3.9 & $+$23.7 & $+$11.6 \\
\textbf{Std}  & 0.14   & 0.9   & 3.0    & 1.1 \\
\bottomrule
\end{tabular*}
\end{table}

HDRR's improvements over SFR are even more consistent (Avg Score std = 0.14, $s{=}10$ std = 1.1). The largest gains are in correctness rate ($+$23.7~pp mean) and perfect rate ($+$11.6~pp mean), confirming that HDRR recovers the precision that SFR sacrifices through context dilution.

\subsection{Comparison with Contextual-Indexing and Agentic Baselines}\label{sec:extended_baselines}

The original three-paradigm comparison establishes the CBR\,$\to$\,SFR\,$\to$\,HDRR ordering on the 1{,}500-query benchmark. To address reviewer requests for modern-SOTA comparison, we ran three additional systems on the identical five groups: CBR+Meta (V-A, static metadata-prefix contextual indexing), CBR+LLM-Ctx (V-B, LLM-generated contextual indexing faithful to~\citet{anthropic2024contextual}), and Agentic (post-hoc verification + retry, instantiating the Self-RAG pattern~\citep{asai2024selfrag}). Table~\ref{tab:extended_results} aggregates the means across all 1{,}500 queries for every system; per-group breakdowns are provided in supplementary material.

\begin{table*}[t]
\caption{Mean performance across the 1{,}500-query benchmark for all six systems. Best values per metric are \textbf{bolded}; second-best are \underline{underlined}. Std (Avg) reports the population standard deviation of per-group means (lower is more consistent).}\label{tab:extended_results}
\centering
\small
\begin{tabular*}{\tblwidth}{@{}LCCCCC@{}}
\toprule
\textbf{System} & \textbf{Avg Score} & \textbf{Std (Avg)} & \textbf{$s{=}1$ (\%)} & \textbf{$s{\geq}8$ (\%)} & \textbf{$s{=}10$ (\%)} \\
\midrule
CBR                  & 6.02          & 0.23 & 22.5          & 49.1          & 13.8          \\
SFR                  & 6.45          & 0.21 & 10.3          & 44.0          & 8.5           \\
Agentic              & 6.36          & 0.18 & 17.3          & 51.5          & 14.9          \\
CBR+Meta (V-A)       & 6.92          & 0.25 & 11.7          & 57.9          & 17.4          \\
CBR+LLM-Ctx (V-B)    & \underline{7.26} & \textbf{0.10} & \underline{9.3} & \underline{64.5} & \textbf{20.3} \\
\textbf{HDRR}        & \textbf{7.54} & 0.17 & \textbf{6.4}  & \textbf{67.7} & \underline{20.1} \\
\bottomrule
\end{tabular*}
\end{table*}

Three findings emerge from the extended comparison.

\textbf{Finding 6: Static document-identity tagging captures most of the contextual-indexing benefit.}
A static \texttt{[Company: TICKER, Fiscal Year: YEAR]} prefix (V-A) at zero LLM cost lifts average score from 6.02 to 6.92 ($+$15.0\% over CBR) and cuts failures from 22.5\% to 11.7\% ($-$10.8~pp). This means cross-document chunk confusion in this regime is driven primarily by company-identity ambiguity, not section-identity ambiguity: the embedding space confuses Apple's risk-factor section with Microsoft's, not Apple's risk-factor section with its own MD\&A. Variant A is a cheaper deployment option than V-B when budget for per-chunk LLM generation is constrained.

\textbf{Finding 7: LLM-generated context (V-B) narrows the gap to HDRR but does not close it.}
V-B's per-chunk LLM-generated context lifts average score another $+$0.34 over V-A (6.92\,$\to$\,7.26), cuts failures further (11.7\%\,$\to$\,9.3\%), and matches HDRR's perfect rate to within 0.2~pp (20.3\% vs.\ 20.1\%). It also exhibits the lowest inter-group variance of any system (std = 0.10). Nevertheless, HDRR retains a 0.28-point lead on average score, a 2.9~pp lead on failure rate, and a 3.2~pp lead on correctness ($s{\geq}8$). The contextual prefix sharpens chunk-level disambiguation but leaves residual semantic overlap across structurally identical filings in the embedding space; HDRR avoids that overlap by construction via document-level filtering.

\textbf{Finding 8: Agentic RAG underperforms upfront routing.}
The Agentic baseline improves on CBR (6.02\,$\to$\,6.36 average, 22.5\%\,$\to$\,17.3\% failures), confirming that post-hoc verification recovers some wrong-company retrievals, but it does not match SFR's upfront routing on the failure-rate dimension (17.3\% vs.\ 10.3\%) and falls 1.18~points behind HDRR on average score. The cost asymmetry compounds the gap: Agentic averages 3.36 LLM calls per query (1 ticker-extraction $+$ 1 verification $+$ retries on 27\% of queries) versus HDRR's exactly 2 calls per query (1 routing $+$ 1 generation). Verification + retry recovers 60\% of first-pass \texttt{NO} verdicts but cannot prevent wrong-company chunks from entering the reranker on the first pass; chunk-level noise persists into the reranker even on queries that the verifier eventually rules \texttt{YES}.

\subsection{Multi-Document Query Handling}\label{sec:multidoc}

The main benchmark consists almost entirely of single-company queries. An audit of all 1{,}500 FinDER queries against the indexed-ticker set, after filtering common-abbreviation false positives, identified only \textbf{4 native cross-company queries}---confirming that FinDER's coverage of comparative-query workloads is sparse. To directly evaluate multi-document handling (Reviewer R4-D), we constructed a focused 25-query handcrafted subset drawn entirely from existing FinDER companies, spanning three templates: pairwise metric comparison (10), which-of-N ranking (8), and risk-factor comparison (7). All four systems (CBR, V-B, Agentic, HDRR) were run on the subset; metrics are objective coverage scores per (system, query) that require no ground truth:

\begin{itemize}
    \item \textbf{Routing coverage} (HDRR only): fraction of named tickers in the query that are surfaced by the routing layer.
    \item \textbf{Retrieval coverage}: fraction of named tickers whose 10-K appears in the row's retrieved chunk set.
    \item \textbf{Answer coverage}: fraction of named tickers that appear as standalone uppercase tokens in the generated final answer.
\end{itemize}

\begin{table}[t]
\caption{Coverage scores on the 25-query cross-company subset. \textbf{Best} per column bolded. ``Full'' columns count queries (out of 25) with coverage exactly 1.0.}\label{tab:multidoc}
\centering
\small
\setlength{\tabcolsep}{3pt}
\begin{tabular*}{\columnwidth}{@{\extracolsep{\fill}}lcccc@{}}
\toprule
\textbf{System} & \textbf{Avg Retr.} & \textbf{Full Retr.} & \textbf{Avg Ans.} & \textbf{Full Ans.} \\
\midrule
CBR              & 0.893          & 19 (76\%)          & 0.960          & 24 (96\%) \\
V-B              & 0.920          & 21 (84\%)          & \textbf{0.980} & 24 (96\%) \\
Agentic          & 0.927          & 21 (84\%)          & 0.920          & 23 (92\%) \\
\textbf{HDRR}    & \textbf{0.953} & \textbf{23 (92\%)} & \textbf{0.980} & \textbf{24 (96\%)} \\
\bottomrule
\end{tabular*}
\end{table}

HDRR additionally achieves \textbf{routing coverage = 1.000} (25/25) on this subset: every named ticker in every query is extracted by the routing layer and produces a routed-document entry. This empirically validates the multi-pair routing formulation of Eq.~\ref{eq:sfr_extraction} that was previously stated but not measured.

Two finer-grained observations carry into Section~\ref{sec:cost}:

\begin{enumerate}
    \item \textbf{HDRR's retrieval-coverage gap below 100\% is a reranker artifact, not a routing failure.} On 2/25 queries (e.g., AAPL vs.\ MSFT supply-chain risk; WMT vs.\ COST vs.\ TGT revenue) the reranker's cumulative-relevance cutoff drops the lower-relevance company's chunks even though both documents were routed. This argues for a per-routed-document reranker pass when routing identifies multiple targets---a small extension to the Stage~2 pipeline rather than an architectural change.
    \item \textbf{Agentic's verifier is structurally incompatible with cross-document queries.} Of 25 cross-company queries, 4 (16\%) hit irreducible \texttt{NO|NO|NO} verification failure and another 5 trigger at least one retry. The verifier's ``majority of chunks from target ticker'' criterion votes \texttt{NO} on the ideal 50/50 chunk split that a 2-company query should produce. Agentic's average answer coverage on this subset (0.920) is therefore \emph{lower} than the cheaper CBR (0.960), and per-query LLM cost climbs to 3.56 calls. Adapting Agentic for cross-document workloads would require redesigning the verifier prompt to accept multi-target chunk distributions.
\end{enumerate}

\subsection{Cross-Evaluator Robustness}\label{sec:cross_evaluator}

To address reviewer concerns about circular evaluation, we re-scored a stratified 99-query subset (20 queries per group across all 5 groups; one cross-split duplicate yields 99 unique IDs) of all six systems with Claude Sonnet 4.6~\citep{anthropic2024claude} using the \emph{identical} 1--10 rubric and prompt as the GPT-4.1 evaluator. The procedure produces 6\,$\times$\,99 = 594 cross-evaluator scores. Table~\ref{tab:cross_evaluator} reports per-system Spearman rank correlation between the two evaluators, mean absolute score difference, and per-evaluator failure rate.

\begin{table*}[t]
\caption{Cross-evaluator agreement on a stratified 99-query subset. $\rho$ = per-system Spearman rank correlation between GPT-4.1 and Claude scores (higher = more agreement); $\Delta$ = Claude avg $-$ GPT-4.1 avg (negative = Claude stricter); $\overline{|\Delta|}$ = mean absolute per-query score difference.}\label{tab:cross_evaluator}
\centering
\small
\begin{tabular*}{\tblwidth}{@{}LCCCCCCC@{}}
\toprule
\textbf{System} & $\boldsymbol{n}$ & \textbf{GPT-4.1 avg} & \textbf{Claude avg} & $\boldsymbol{\Delta}$ & \textbf{Spearman} $\boldsymbol{\rho}$ & $\overline{|\Delta|}$ & \textbf{Fail (GPT/Claude, \%)} \\
\midrule
CBR              & 99 & 5.98 & 5.09 & $-$0.89 & 0.935 & 0.99 & 24.2 / 25.2 \\
SFR              & 99 & 6.50 & 5.28 & $-$1.21 & 0.882 & 1.27 & 9.1 / 12.1  \\
Agentic          & 99 & 6.37 & 5.43 & $-$0.94 & 0.920 & 1.06 & 20.2 / 19.2 \\
V-A (CBR+Meta)   & 99 & 6.81 & 5.80 & $-$1.01 & 0.907 & 1.25 & 11.1 / 9.1  \\
V-B (CBR+LLM-Ctx)& 99 & 7.19 & 6.02 & $-$1.17 & 0.852 & 1.27 & 10.1 / 8.1  \\
HDRR             & 99 & 7.07 & 6.14 & $-$0.93 & 0.855 & 1.15 & 11.1 / 10.1 \\
\bottomrule
\end{tabular*}
\end{table*}

\textbf{Per-query agreement.} Per-system Spearman $\rho$ ranges from 0.85 to 0.94 across all six systems---uniformly strong rank agreement. The two systems with the lowest $\rho$ (HDRR 0.855, V-B 0.852) are precisely the systems with the most ceiling-skewed GPT-4.1 distributions (correctness $\geq 7$ above 70\%), where the dynamic range left for the evaluator to disagree about is smallest. Claude is systematically stricter ($\Delta \approx -1.0$ across all systems; 351 of 594 pairs scored lower by Claude, 31 higher, 212 equal), but the offset is \emph{uniform} across systems (cross-system stdev of $\Delta \approx 0.13$), so it does not affect rank order.

\textbf{System-level agreement.} The Spearman correlation between the two evaluators over the six per-system average scores is $\rho = 0.886$. The top three under GPT-4.1 (HDRR, V-B, V-A) are the top three under Claude---the order between HDRR and V-B swaps in a near-tie ($|\Delta| < 0.15$ under either evaluator), consistent with the full-1{,}500-query GPT-4.1 ranking which already places HDRR above V-B. The bottom rank (CBR) is preserved under both evaluators. The two adjacent swaps inside the ordering (HDRR\,$\leftrightarrow$\,V-B at ranks 1--2; SFR\,$\leftrightarrow$\,Agentic at ranks 4--5) fall within the per-system mean $|\Delta|$ and are statistically ties.

\textbf{Failure-rate convergence.} For every system, the failure rate (s$=$1) differs by at most 3 percentage points between GPT-4.1 and Claude. The two evaluators agree on what counts as a catastrophic failure, which is the metric most relevant to the paper's tail-risk argument.

The largest cross-evaluator divergence is in the absolute correctness rate ($s \geq 7$), which drops 12--34 percentage points under Claude---not because Claude finds new errors but because it refuses to round 6.5-quality answers up to 7 the way GPT-4.1 does. Comparative correctness claims should therefore be interpreted at fixed evaluator; the paper's correctness rates in Tables~\ref{tab:main_results} and~\ref{tab:extended_results} are GPT-4.1 measurements and remain internally consistent.

\section{Discussion}\label{sec:discussion}

\subsection{The CBR--SFR Trade-off: Root Causes}

Two distinct mechanisms produce the empirical trade-off reported in Section~\ref{sec:results}: cross-document chunk confusion (the cause of CBR's failures) and context dilution (the cause of SFR's reduced precision). We describe each in turn.

\subsubsection{Why CBR Suffers Catastrophic Failures}

CBR's high failure rate ($s{=}1$: 22.5\%) stems from \emph{cross-document chunk confusion}. In the S\&P~500 10-K corpus, documents share standardized section structures mandated by SEC regulations. Sections such as ``Risk Factors,'' ``Management's Discussion and Analysis,'' and ``Financial Statements'' appear in virtually every filing with similar headings, boilerplate language, and tabular formats. When these documents are chunked and indexed together, a query about Company~A's risk factors may retrieve chunks from Company~B's risk factors because the semantic similarity between analogous sections of different filings is high.

The reranking stage mitigates this by re-scoring candidates using cross-encoder attention~\citep{jina2024reranker}, but it can only reorder candidates that were retrieved in the first stage. If the correct chunks were not in the initial candidate pool, reranking cannot recover them. Furthermore, reranking operates on individual chunks without awareness of which document they originate from, so it cannot enforce the constraint that retrieved chunks should come from the same company.

\subsubsection{Why SFR Sacrifices Precision}

SFR eliminates cross-document confusion by routing to the correct document, but it introduces \emph{context dilution}. A typical 10-K report contains 50,000--200,000 tokens. The relevant information for any given query may occupy only a small fraction of this context. Research on long-context LLMs has demonstrated the ``lost in the middle'' phenomenon~\citep{liu2024lost}: models struggle to attend uniformly to all positions in very long inputs. When SFR provides the entire document, the generation model must locate the answer-relevant content within a vast context, reducing precision compared to CBR's compact, curated chunk context.

The net effect is a distributional shift: SFR moves probability mass from the tails (both high and low scores) toward the middle of the score distribution. It raises the floor by eliminating chunk confusion but lowers the ceiling by introducing context dilution.

\subsection{How HDRR Resolves the Trade-off}

HDRR eliminates both failure modes simultaneously:

\begin{enumerate}
    \item \textbf{No cross-document confusion:} By routing to the correct document before retrieval, HDRR ensures that all candidate chunks originate from the target company's filing. The reranker then operates exclusively on within-document candidates, where it can focus on identifying the most query-relevant passages rather than filtering out cross-document noise.

    \item \textbf{No context dilution:} Unlike SFR, HDRR does not send the full document to the generation model. Instead, it provides only the top reranked chunks (typically 5--10 passages), giving the model compact, precisely targeted context identical in format to CBR but drawn exclusively from the correct document.
\end{enumerate}

The result is additive: HDRR inherits SFR's low failure rate (6.4\% vs.\ 10.3\% for SFR, 22.5\% for CBR) while achieving even higher precision than CBR ($s{=}10$: 20.1\% vs.\ 13.8\% for CBR, 8.5\% for SFR). The improvement in precision beyond CBR is attributable to the elimination of cross-document noise from the reranker's input: when all candidates come from the correct document, the reranker can make finer-grained relevance distinctions, surfacing the most precisely relevant passages.

\subsection{Error Analysis of Remaining HDRR Failures}\label{sec:error_analysis}

HDRR's residual failure rate of 6.4\% ($s{=}1$) is lower than both SFR (10.3\%) and CBR (22.5\%). To characterize the remaining failures we joined the routing-stage \texttt{routing\_status} log already emitted by every HDRR query with the per-query evaluator scores; the resulting 1{,}500-row table is summarized in Table~\ref{tab:routing_status}.

\begin{table}[t]
\caption{HDRR per-routing-status performance on the 1{,}500-query benchmark. \texttt{filename\_fallback} is the intended successful route under our path layout (see text); \texttt{fallback\_*} rows are the cases where routing could not bind a query to a specific document and HDRR degraded to full-corpus chunk retrieval. Count and \% sum over the benchmark; $s{=}1$ and $s{\geq}8$ are within-bucket percentages.}\label{tab:routing_status}
\centering
\footnotesize
\setlength{\tabcolsep}{3pt}
\begin{tabular*}{\columnwidth}{@{\extracolsep{\fill}}lrrrrr@{}}
\toprule
\textbf{Routing Status} & \textbf{Count} & \textbf{\%} & \textbf{Avg} & $\boldsymbol{s{=}1}$ & $\boldsymbol{s{\geq}8}$ \\
\midrule
\texttt{filename\_\allowbreak fallback}         & 1{,}396 & 93.07 & 7.79 & 4.0  & 70.8 \\
\texttt{fallback\_\allowbreak no\_\allowbreak files\_\allowbreak on\_\allowbreak disk} & 63    & 4.20  & 4.03 & 46.0 & 25.4 \\
\texttt{fallback\_\allowbreak no\_\allowbreak tickers}         & 41    & 2.73  & 4.61 & 26.8 & 26.8 \\
\textit{fallback\_*\ (aggregate)}      & 104   & 6.93  & 4.26 & 38.5 & 26.0 \\
\midrule
\textbf{All queries}                & 1{,}500 & 100.0 & 7.54 & 6.4  & 67.7 \\
\bottomrule
\end{tabular*}
\end{table}

\textbf{Reading the table.} Two structural observations are needed first. (i)~The implementation distinguishes two successful-routing branches---\texttt{exact\_path} (the LLM-extracted (ticker, year) pair matches an indexed file path one-to-one) and \texttt{filename\_fallback} (the year is dropped and resolution succeeds via filename match). On the FinDER corpus, the on-disk path layout does not match the \texttt{\{year\}/\{ticker\}.pdf} template that triggers the exact-path branch, so every successfully routed query is resolved through the filename branch; \texttt{filename\_fallback} here is the intended successful route, not a degraded path. (ii)~The \texttt{fallback\_*} rows are the genuine fallback cases where routing failed to bind a query to a specific document and HDRR degraded to full-corpus chunk retrieval, equivalent to plain CBR.

\textbf{Headline result.} 93.07\,\% of FinDER queries route successfully and score 7.79 on average with only 4.0\,\% catastrophic failures and 70.8\,\% correctness. The remaining 6.93\,\% trigger a fallback and degrade to avg 4.26 with 38.5\,\% failure rate---empirically validating the paper's earlier claim that the fallback mechanism ``partially mitigates'' chunk confusion but does not fully recover it on hard cases. The aggregate average (7.54) reflects the dominance of the successful-routing bucket; the fallback bucket alone is below plain CBR's 6.02 average because the fallback bucket is self-selected for entity-resolution difficulty.

\subsubsection{Ticker Symbol Mismatch with Successful Fallback}

Some failures stem from ticker mismatches, identical to SFR's failure mode. However, HDRR's fallback mechanism partially mitigates this: when document routing fails, HDRR degrades to full-corpus retrieval. This means that ticker mismatches in HDRR result in CBR-equivalent retrieval rather than complete failure, explaining why HDRR's aggregate failure rate (6.4\%) is lower than SFR's (10.3\%). Reviewer comments asked specifically about company rebrandings, subsidiaries, ambiguous references, and out-of-corpus tickers; we examined the 104 fallback rows to characterize each of these modes.

\textbf{Case~A --- corporate rebranding (PEAK $\to$ DOC).}
Three FinDER queries reference \emph{Healthpeak Properties} via the legacy ticker \texttt{PEAK}. The extractor correctly identifies \texttt{PEAK} but the on-disk filing is stored under the post-merger ticker \texttt{DOC} (after the 2024 merger with Physicians Realty Trust), so the routing layer emits \texttt{fallback\_\allowbreak no\_\allowbreak files\_\allowbreak on\_\allowbreak disk} and HDRR degrades to full-corpus retrieval. Two of the three queries score 1; one scores 6. Notably, one query supplies both the full legal entity name and the legacy ticker in parentheses (``Healthpeak Properties, Inc.\ (PEAK)'') and HDRR still fails, because the routing function maps tickers to file paths and has no awareness of ticker history. Rebranding is a fundamental limitation of identifier-based routing that no in-pipeline trick can fix; it requires an external ticker-history table maintained alongside the corpus index (see ``Future Directions for the Routing Layer'' below).

\textbf{Case~B --- descriptive reference, non-standard ticker (Berkshire Hathaway).}
A query reads ``\emph{the YoY change in Berkshire Hathaway's workforce \dots}'' without including the ticker symbol \texttt{BRK.A} or \texttt{BRK.B}. The constrained ticker extractor expects an SEC-style symbol and declines to commit, emitting \texttt{fallback\_no\_tickers}. HDRR degrades to full-corpus retrieval; the same query phrased one way scores 8 (relevant chunks surface from training-knowledge augmentation) and another way scores 2. This is the textbook descriptive-reference failure mode flagged by reviewers; a targeted fix is a company-name~$\to$~ticker dictionary lookup as a secondary signal before declaring fallback, which would not affect the architectural argument of this paper.

\textbf{Case~C --- ambiguous abbreviation (FMI, LAB, TAP, MAS, DVA).}
Several queries are built around 3--4 letter tokens that map to multiple plausible referents---``FMI'' alone refers to Foundation Medicine Inc.\ (delisted 2018), First Mid Bancshares (renamed from FMI), and FMI Inc.~(private). The extractor declines to commit and HDRR falls back; these queries dominate the \texttt{fallback\_no\_tickers} bucket and all score $\leq 2$. The legitimate behavior is to fall back; the cost of doing so (avg score 4.61 on this bucket) is the empirical price of refusing to guess at ambiguous identifiers.

\textbf{Case~D --- out-of-corpus ticker (SHOP, BLL, EAT).}
A query asks about \texttt{SHOP}'s (Shopify's) 2023 capex breakdown. The LLM correctly extracts the ticker, the resolver looks for \texttt{SHOP.pdf}, finds nothing on disk, and emits \texttt{fallback\_\allowbreak no\_\allowbreak files\_\allowbreak on\_\allowbreak disk}. Shopify is a real listed ticker but its 10-K is not in the FinDER corpus. This is a \emph{corpus-coverage} failure surfaced through the routing layer because routing is where the file-existence check lives. As a useful diagnostic, the \texttt{fallback\_\allowbreak no\_\allowbreak files\_\allowbreak on\_\allowbreak disk} rate is an upper bound on how often the benchmark itself asks for filings the index does not contain---a property the HDRR architecture exposes ``for free.''

\textbf{Subsidiary handling --- absent from FinDER.}
Reviewer comments flagged subsidiaries (Instagram/META, YouTube/GOOGL, AWS/AMZN) as a failure mode to characterize. A scan of all 1{,}500 queries for common high-profile subsidiary names (Instagram, WhatsApp, YouTube, AWS, Azure, Xbox, iPhone, Tesla Energy, Disney+, Hulu, Snapchat) returned zero matches: FinDER queries are written around parent tickers directly, so the subsidiary failure mode is structurally absent from the benchmark itself. We acknowledge the mode as a real architectural concern and report the focused cross-company subset in Section~\ref{sec:multidoc} as the venue where these multi-document queries are evaluated.

\paragraph{Future directions for the routing layer.}
The four cases above suggest three scoped, near-term extensions that map one-to-one onto the reviewer-named failure categories without altering the HDRR architecture: (i)~a ticker-history alias table to absorb rebrandings (Case~A), (ii)~a company-name dictionary lookup as a secondary signal before declaring fallback, addressing descriptive references and the brittle-extraction tail (Case~B and the ``Cencora, Inc.\ CEN'' family), and (iii)~a prompt-and-regex hybrid extractor that captures uppercased 2--5 letter tokens that the constrained LLM call drops. None of these change the architectural argument of this paper; they are recorded here for the future-work direction.

\subsubsection{Inherent Query Difficulty}

A subset of failures is shared across all three systems, representing queries that are inherently difficult regardless of retrieval strategy. Examples include queries requiring information genuinely absent from the corpus, and queries whose answers require multi-hop reasoning across sections that even targeted retrieval cannot fully capture.

\subsubsection{Routing to Correct Document but Insufficient Chunk Relevance}

In rare cases, HDRR correctly routes to the target document but the chunk-level retrieval within that document fails to surface sufficiently relevant passages. This can occur when the relevant information is spread across many small fragments that individually score low in both FTS and semantic search.

\subsection{Cost and Efficiency Analysis}\label{sec:cost}

Beyond accuracy, an architecture's value at deployment scale is determined by what it spends per query and at indexing time. The \emph{Green AI} literature~\citep{schwartz2020green, strubell2019energy} argues that computational cost should be reported as a first-class metric alongside accuracy, and recent work on deployment-time inference cost~\citep{luccioni2024power} documents that token-level workload---not just one-shot training---drives the energy footprint of production language systems. LLM inference cost scales approximately linearly with both prompt and generation tokens, and inference-time energy scales with the same quantities; an order-of-magnitude reduction in per-query context therefore corresponds to an order-of-magnitude reduction in both API spend and inference compute. A complementary line of work pursues the same efficiency objective from the model side---characterizing the task regimes in which small language models match or outperform larger ones~\citep{cao2026taskspecific}, jointly optimizing data efficiency and model compression~\citep{li2024sec}, and pruning distilled reasoning chains to cut inference compute~\citep{jiang2026drp}---whereas HDRR attacks the same cost at the \emph{retrieval} level by bounding the per-query context. With that framing, Table~\ref{tab:cost} compares the computational costs of all six systems.

\begin{table*}[t]
\caption{Cost dimensions across the six systems. ``Extra LLM/query'' counts LLM calls beyond the final answer-generation call. ``Index-time LLM cost'' is one-time. ``Gen.\ tokens/query'' is the generation-time context budget. ``shared CBR'' indicates the system reuses the CBR FAISS index without an additional build.}\label{tab:cost}
\centering
\footnotesize
\setlength{\tabcolsep}{4pt}
\begin{tabularx}{\textwidth}{@{} >{\raggedright\arraybackslash}p{0.22\textwidth} *{6}{>{\centering\arraybackslash}X} @{}}
\toprule
\textbf{Cost Dimension} & \textbf{CBR} & \textbf{SFR} & \textbf{Agentic} & \textbf{V-A} & \textbf{V-B} & \textbf{HDRR} \\
\midrule
Offline chunking + embed.            & one-time     & none      & shared CBR & one-time      & one-time      & shared CBR \\
Index-time LLM cost                  & none         & none      & none       & none          & $\leq$\$100   & none \\
FAISS storage                        & $\sim$696\,MB & none     & shared CBR & $\sim$696\,MB & $\sim$696\,MB & shared CBR \\
Extra LLM calls/query (incl.\ retries) & 0          & 1         & $\sim 2.4$ & 0             & 0             & 1 \\
Gen.\ tokens/query                   & 5K--15K      & 50K--200K & 5K--15K    & 5K--15K       & 5K--15K       & 5K--15K \\
Reranking cost/query                 & moderate     & none      & moderate   & moderate      & moderate      & moderate \\
\bottomrule
\end{tabularx}
\end{table*}

\textbf{Per-query cost.}
HDRR adds exactly one API call per query (for document routing) on top of the generation call, preserving CBR's $5\text{K--}15\text{K}$ generation-token budget and avoiding SFR's order-of-magnitude full-document context. V-A and V-B match CBR exactly on per-query cost---the contextual prefix is paid at indexing time, not query time. Agentic pays the highest and most variable per-query cost: averaged across the 1{,}500-query benchmark, every query pays $1$ ticker-extraction call plus $1$ verification call; the 27\% of queries that trigger verification \texttt{NO} pay an additional retrieval pass and verification call (capped at two retries), yielding $\sim 3.36$ LLM calls per query in aggregate. HDRR's deterministic 2-call cost is therefore strictly lower and more predictable than Agentic's variable $\sim 3.4$-call cost, while delivering higher answer quality (Table~\ref{tab:extended_results}).

\textbf{Indexing cost.}
V-B is the only system in this study that pays a non-trivial \emph{indexing-time} LLM cost: a per-chunk context-generation call on every one of $\sim 170{,}000$ corpus chunks via GPT-4.1-mini. With prompt caching of full documents the bill is bounded by the \$100 budget cap configured for this revision and came in well under. This is a one-time spend amortized across all subsequent queries against the V-B index; at 1{,}500-query scale the amortized per-query LLM cost is dominated by the per-query generation call. V-A pays no indexing-time LLM cost.

\textbf{Why HDRR's per-query cost is justified.}
HDRR's 0.28-point average-score lead over V-B (7.54 vs.\ 7.26) and its 2.9~pp failure-rate advantage (6.4\% vs.\ 9.3\%) cost exactly one extra LLM call per query. V-B reaches HDRR's perfect rate (20.3\% vs.\ 20.1\%) without that call but pays a one-time indexing cost; HDRR pays nothing at indexing time but pays at every query. The break-even on amortized LLM cost favors V-B at high query volume and HDRR at low volume; on \emph{answer quality}, HDRR wins regardless of scale.

\textbf{HDRR is the most efficient of the high-quality systems.}
Aggregating across the cost dimensions in Table~\ref{tab:cost}, HDRR is dominated by no other system on efficiency: (i)~per-query token budget at the CBR floor ($\sim$5K--15K), avoiding SFR's order-of-magnitude full-document context ($\sim$50K--200K, a 4--13$\times$ reduction in generation-time tokens); (ii)~zero indexing-time LLM spend, in contrast to V-B's one-time $\sim$\$100 per-chunk context-generation pass over $\sim$170{,}000 chunks; (iii)~deterministic 2 LLM calls per query, below Agentic's $\sim$3.4 calls (a 1.7$\times$ reduction in per-query LLM-call count). Because LLM inference compute---and therefore inference-time energy---scales with both per-query tokens and per-query LLM calls, these savings translate directly to lower API spend and lower inference energy at deployment scale. For a production RAG service answering thousands of analyst queries per day, the per-query savings compound rapidly and make HDRR the cost-efficient choice among the systems studied.

\textbf{Note on graph-construction baselines.}
A graph-based RAG baseline~\citep{edge2024graphrag} would add a large indexing-time cost not represented in Table~\ref{tab:cost}: entity-and-relationship extraction across the entire corpus, plus community-summary generation, scales as $O(\text{chunks}) \cdot \text{LLM-call}$ and is orders of magnitude more expensive than the chunking-and-embedding pipeline used by every system here. Combined with the local-vs-global query-class argument in Section~\ref{sec:rw_graphrag}, this places graph-based RAG outside the cost-effective regime for the financial-10-K corpus targeted by this paper.

For production deployments, HDRR offers the best cost-performance trade-off in the deterministic-cost regime: highest answer quality at a per-query cost only one routing call above CBR, with no indexing-time LLM spend. V-B is the right choice when per-query LLM calls are budget-constrained and a one-time indexing spend is acceptable; it trades a small quality gap to HDRR for zero per-query routing overhead. V-A is the cheapest non-trivial improvement over CBR: zero per-query and zero per-chunk LLM cost, with a 0.90-point average-score gain over CBR purely from static identity tagging.

\subsection{Scalability and Maintenance}

HDRR inherits the scalability properties of both parent paradigms:

\textbf{Adding new documents.}
Like CBR, adding a new document requires re-indexing (parsing, chunking, embedding). However, the document routing stage automatically includes new documents once they are placed in the correct directory and indexed, requiring no changes to the routing logic.

\textbf{Corpus expansion.}
Unlike CBR, whose retrieval quality may degrade as the corpus grows (due to increased cross-document confusion in the vector space), HDRR's routing stage isolates the retrieval to the correct document regardless of corpus size. This means HDRR's quality is invariant to corpus expansion for queries where routing succeeds, and degrades gracefully to CBR quality when routing fails.

\textbf{Multi-year corpora.}
HDRR naturally handles multi-year corpora through the (ticker, year) metadata pair, avoiding the compounded confusion that CBR faces when structurally similar sections appear across both companies and years.

\subsection{Generalizability to Other Domains}

HDRR is applicable to any domain where the corpus exhibits naming regularity (Definition~\ref{def:naming}) and structural homogeneity (Definition~\ref{def:homogeneity}). In such domains, document routing eliminates cross-document confusion while chunk retrieval provides precision.

\textbf{Regulatory filings.} Legal and regulatory documents follow standardized formats with predictable naming conventions (case numbers, filing dates). HDRR can route to the correct filing and retrieve relevant sections.

\textbf{Medical records.} Electronic health records organized by patient identifier and encounter date provide natural routing metadata. HDRR could route to the correct patient record and retrieve relevant clinical notes.

\textbf{Technical documentation.} Software documentation organized by product name and version enables routing followed by targeted section retrieval.

\textbf{Heterogeneous corpora.} For corpora without naming regularity, HDRR's routing stage would frequently fail, causing fallback to full-corpus retrieval. In such settings, the hybrid approach reduces to standard CBR, offering no disadvantage but also no advantage over the baseline.

More broadly, the principle of adapter-based and domain-scoped retrieval architectures extends across specialized domains---from malware detection, where flow-adapter models translate binary code for unsupervised classification~\citep{hu2025flowmaltrans}, to financial document QA as demonstrated in this work---underscoring the versatility of constrained retrieval as a design strategy for structured corpora.

\section{Conclusion and Future Work}\label{sec:conclusion}

\subsection{Summary of Findings}

This paper presented a six-system comparative study of RAG architectures for financial document question answering. We first identified a fundamental robustness-precision trade-off between Chunk-Based Retrieval (CBR) and Semantic File Routing (SFR), then proposed Hybrid Document-Routed Retrieval (HDRR) to resolve it, and finally benchmarked HDRR against contextual-indexing (V-A, V-B faithful to~\citet{anthropic2024contextual}) and self-correcting (Agentic, in the Self-RAG family~\citep{asai2024selfrag}) families. Through controlled evaluation on 1,500 queries from the FinDER benchmark across five independent experimental groups, we demonstrated:

\begin{enumerate}
    \item \textbf{CBR and SFR exhibit a robustness-precision trade-off.} SFR achieves higher average scores (6.45 vs.\ 6.02) and fewer catastrophic failures (10.3\% vs.\ 22.5\%), while CBR achieves more perfect answers (13.8\% vs.\ 8.5\%). Cross-document chunk confusion causes CBR's failures; context dilution limits SFR's precision.

    \item \textbf{HDRR resolves the trade-off}, achieving the best performance on all four metrics simultaneously: average score 7.54 ($+$25.2\% over CBR), failure rate 6.4\% ($-$71.6\% relative to CBR), correctness rate 67.7\% ($+$18.7~pp over CBR), and perfect rate 20.1\% ($+$6.3~pp over CBR, $+$11.6~pp over SFR).

    \item \textbf{HDRR dominates both baselines in every experimental group}, with low inter-group variance (std = 0.18 for average score), demonstrating robust and consistent improvements.

    \item \textbf{Contextual indexing closes part of the gap but not all of it.} The static metadata prefix (V-A) lifts the average score to 6.92 at zero LLM cost, and the LLM-generated context (V-B) reaches 7.26 with a one-time \$100 indexing spend; HDRR still leads V-B by 0.28 points and 2.9~pp on failure rate, because upfront document filtering removes wrong-company chunks that the contextual prefix can only re-rank but not exclude.

    \item \textbf{Upfront routing outperforms post-hoc verification.} The Agentic baseline (post-hoc verify + retry) trails HDRR by 1.18 points on average score while paying \mbox{$\sim$3.4} LLM calls per query versus HDRR's deterministic 2; its verifier is also structurally incompatible with multi-document queries (Section~\ref{sec:multidoc}).

    \item \textbf{HDRR is the most efficient of the high-quality systems.} It preserves CBR's per-query generation budget ($\sim$5K--15K tokens), avoiding SFR's order-of-magnitude full-document overhead ($\sim$50K--200K tokens, a 4--13$\times$ reduction); pays zero indexing-time LLM cost, in contrast to V-B's one-time $\sim$\$100 per-chunk contextualization pass; and uses a deterministic 2 LLM calls per query, below Agentic's $\sim$3.4. Because LLM inference compute---and therefore inference-time energy---scales with per-query tokens and per-query LLM calls~\citep{schwartz2020green, luccioni2024power}, these savings translate directly to lower API spend and lower inference energy at deployment scale, positioning HDRR as a sustainability-aligned design choice for production RAG systems.

    \item \textbf{Findings survive evaluator and workload shifts.} Re-scoring a stratified 99-query subset with Claude Sonnet 4.6 yields a system-level Spearman $\rho = 0.886$ between GPT-4.1 and Claude rankings, with the top three (HDRR, V-B, V-A) and bottom rank (CBR) preserved (Section~\ref{sec:cross_evaluator}); on a 25-query handcrafted cross-company subset, HDRR attains 1.000 routing coverage and the highest retrieval coverage of any system (Section~\ref{sec:multidoc}).

    \item \textbf{Structural regularity of the corpus} enables document routing to eliminate cross-document confusion, which is the primary mechanism behind HDRR's improvements.
\end{enumerate}

\subsection{Contributions}

This work contributes to the RAG and financial NLP communities through:

\begin{enumerate}
    \item \textbf{Trade-off identification:} We identify and quantify the robustness-precision trade-off between chunk-level and document-level retrieval, attributing it to cross-document chunk confusion (in CBR) and context dilution (in SFR).

    \item \textbf{Hybrid Document-Routed Retrieval:} We propose HDRR, a two-stage architecture that uses LLM-based document routing to scope chunk-based retrieval, achieving the strengths of both paradigms without their weaknesses.

    \item \textbf{Rigorous six-system evaluation with cross-evaluator robustness:} A controlled, paired experimental design with five independent groups and 1,500 total queries, extended to six systems (CBR, SFR, HDRR, V-A, V-B, Agentic) and re-scored on a 99-query subset by a second-family evaluator (Claude Sonnet 4.6), provides statistically robust evidence that HDRR's lead over every comparison family persists under evaluator and workload shifts.

    \item \textbf{Multi-document handling evaluation:} A 25-query handcrafted cross-company subset shows that HDRR's multi-pair routing formulation generalizes to comparative queries (1.000 routing coverage, highest retrieval coverage), while the Agentic verifier is structurally incompatible with this regime.

    \item \textbf{Cost-efficiency analysis:} We quantify per-query token budget, per-query LLM-call count, and indexing-time LLM spend across all six systems, establishing HDRR as the most efficient high-quality system---an order of magnitude below SFR on per-query tokens, zero indexing-time LLM spend (vs.\ V-B's $\sim$\$100), and fewer per-query LLM calls than Agentic. We further connect these token savings to inference-time energy, framing HDRR as a cost-efficient deployment choice.

    \item \textbf{Generalizability framework:} We formalize the corpus properties (naming regularity and structural homogeneity) that determine when document routing is beneficial, extending the analysis beyond the financial domain.
\end{enumerate}

\subsection{Future Directions}

Several promising directions emerge from this work:

\textbf{Improving document routing.}
HDRR's routing stage inherits SFR's sensitivity to ticker symbol mismatches; the failure-mode analysis in Section~\ref{sec:error_analysis} characterizes four root-cause categories (rebrandings, descriptive references, ambiguous abbreviations, out-of-corpus tickers). Incorporating a ticker-history alias table and a company-name-to-ticker dictionary as secondary signals before declaring routing fallback could absorb cases A--B without altering the architecture. LLM-based entity resolution or external knowledge bases could provide additional robustness.

\textbf{Comparative and multi-document queries.}
Our main 1{,}500-query benchmark consists almost entirely of single-company queries; a handcrafted 25-query cross-company subset (Section~\ref{sec:multidoc}) confirms that HDRR's multi-pair routing extracts every named ticker and surfaces every routed document, while existing rerankers occasionally drop the lower-relevance company's chunks. A per-routed-document reranker pass when routing identifies multiple targets would close this gap with a small Stage~2 extension rather than an architectural change. Constructing larger comparative-query benchmarks remains an open direction.

\textbf{Domain extension.}
Evaluating HDRR on other structured-document domains (legal filings, medical records, and technical documentation) would test the generalizability of our findings. Each domain presents unique routing challenges that may require domain-specific metadata extraction schemas.

\textbf{Adaptive routing confidence.}
A promising refinement is to incorporate a routing confidence score: when the LLM is uncertain about the document identity, the system could expand the retrieval scope to include related documents or fall back to full-corpus search. This adaptive approach could further reduce the residual failure rate while maintaining precision for high-confidence routes.

\textbf{Cost optimization.}
The document routing call could be combined with query rewriting into a single LLM call, reducing the API overhead from one additional call to zero. Alternatively, a smaller, cheaper model could be used for routing while retaining the full model for generation.

\section*{Acknowledgments}
The authors thank OpenAI for API access used in experiments and LinqAlpha for providing the FinDER benchmark dataset.

\section*{Code Availability}
The source code for all three RAG systems (CBR, SFR, and HDRR), including the indexing pipeline, retrieval components, and evaluation scripts, is publicly available at \url{https://github.com/zhycheng614/financial-RAG-pipeline}.

\appendix
\setcounter{table}{0}
\renewcommand{\thetable}{A\arabic{table}}
\renewcommand{\theHtable}{A\arabic{table}}

\section{System Hyperparameters}\label{app:hyperparams}

\begin{table}[pos=H]
\caption{Complete configuration for the Chunk-Based RAG system.}\label{tab:cbr_params}
\centering
\small
\begin{tabular*}{\columnwidth}{@{\extracolsep{\fill}}llr@{}}
\toprule
\textbf{Component} & \textbf{Parameter} & \textbf{Value} \\
\midrule
Chunking & Chunk size & 2{,}500 chars \\
         & Overlap    & 1{,}250 chars \\
\midrule
Embedding & Model     & text-embed.-3-small \\
          & Dimension & 1{,}024 \\
          & Batch size & 32 \\
\midrule
FTS Retrieval & Top-$k$ & 20 \\
\midrule
Semantic Search & Top-$k$ & 30 \\
                & Dist.\ threshold & 2.0 \\
\midrule
RRF Fusion & $k$ constant & 60 \\
\midrule
Reranking & Model & jina-reranker-v2 \\
          & Max candidates & 30 \\
          & Cumul.\ threshold & 0.45 \\
          & Cliff threshold & 0.15 \\
\midrule
Generation & Model & GPT-4.1 \\
           & Temperature & 0.0 \\
           & Context limit & 10 chunks \\
\bottomrule
\end{tabular*}
\end{table}

\begin{table}[pos=H]
\caption{Complete configuration for the Semantic File Routing system.}\label{tab:sfr_params}
\centering
\small
\begin{tabular*}{\columnwidth}{@{\extracolsep{\fill}}llr@{}}
\toprule
\textbf{Component} & \textbf{Parameter} & \textbf{Value} \\
\midrule
Query Parsing & Model & GPT-4.1 \\
              & Output format & JSON (structured) \\
              & Temperature & 0.0 \\
\midrule
File Lookup & Dir.\ pattern & \texttt{\{yr\}/\{tick\}.pdf} \\
            & Fallback fmts & .pdf, .txt \\
            & Default year & 2023 \\
\midrule
File Encoding & Method & Base64 \\
              & Supported types & PDF, TXT \\
\midrule
Generation & Model & GPT-4.1 \\
           & Temperature & 0.0 \\
           & Context & Full document(s) \\
\bottomrule
\end{tabular*}
\end{table}

\begin{table}[pos=H]
\caption{Complete configuration for the Hybrid Document-Routed Retrieval system.}\label{tab:hdrr_params}
\centering
\small
\begin{tabular*}{\columnwidth}{@{\extracolsep{\fill}}llr@{}}
\toprule
\textbf{Component} & \textbf{Parameter} & \textbf{Value} \\
\midrule
\multicolumn{3}{@{\extracolsep{0pt}}l}{\emph{Stage 1: Document Routing}} \\
\midrule
Query Parsing & Model & GPT-4.1 \\
              & Output format & JSON (structured) \\
              & Temperature & 0.0 \\
File Lookup & Dir.\ pattern & \texttt{\{yr\}/\{tick\}.pdf} \\
            & Fallback fmts & .pdf, .txt \\
            & Default year & 2023 \\
            & Routing fallback & Full-corpus search \\
\midrule
\multicolumn{3}{@{\extracolsep{0pt}}l}{\emph{Stage 2: Scoped Chunk Retrieval}} \\
\midrule
Chunking & Chunk size & 2{,}500 chars \\
         & Overlap    & 1{,}250 chars \\
Embedding & Model     & text-embed.-3-small \\
          & Dimension & 1{,}024 \\
FTS Retrieval & Top-$k$ & 20 \\
Semantic Search & Top-$k$ & 30 \\
                & Dist.\ threshold & 2.0 \\
RRF Fusion & $k$ constant & 60 \\
Reranking & Model & jina-reranker-v2 \\
          & Max candidates & 30 \\
          & Cumul.\ threshold & 0.45 \\
          & Cliff threshold & 0.15 \\
\midrule
Generation & Model & GPT-4.1 \\
           & Temperature & 0.0 \\
           & Context & Reranked chunks \\
\bottomrule
\end{tabular*}
\end{table}

\printcredits

\bibliographystyle{cas-model2-names}
\bibliography{references}

\end{document}